\documentclass{article}
\PassOptionsToPackage{numbers,compress}{natbib}
\usepackage[preprint]{neurips_2026}
\usepackage{lineno}

\usepackage[utf8]{inputenc} 

\usepackage{graphicx}

\usepackage[T1]{fontenc}

\usepackage[hidelinks,colorlinks]{hyperref}       
\usepackage{url}            
\usepackage{booktabs}       
\usepackage{amsfonts}       
\usepackage{nicefrac}       
\usepackage{microtype}      
\usepackage{pgf}

\IfFileExists{headers/config/showoverfull.config}{
	\overfullrule=1cm
}{
}

\usepackage{marginnote}

\usepackage[backgroundcolor=none,linecolor=red,textsize=footnotesize]{todonotes}

\usepackage{etoolbox}

\newbool{includeappendix}
\setbool{includeappendix}{true} 
\IfFileExists{headers/config/noappendix.config}{
	\setbool{includeappendix}{false}
}{}

\newif\ifincludeappendixx
\ifbool{includeappendix}{
	\includeappendixxtrue
}{
	\includeappendixxfalse
}

\usepackage{xr} 
\usepackage{filecontents}

\ifbool{includeappendix}{}{
	\input{appendix-labels-loader}

	\externaldocument{appendix-labels}
}

\usepackage{xspace}

\newcommand{\eg}{e.g., }
\newcommand{\ie}{i.e., }

\usepackage{acro} 

\usepackage{listings}

\usepackage{textcomp}

\usepackage{xcolor}

\usepackage[scaled=0.8]{beramono}

\definecolor{ckeyword}{HTML}{7F0055}
\definecolor{ccomment}{HTML}{3F7F5F}
\definecolor{cstring}{HTML}{2A0099}

\lstdefinestyle{numbers}{
	numbers=left,
	framexleftmargin=20pt,
	numberstyle=\tiny,
	firstnumber=auto,
	numbersep=1em,
	xleftmargin=2em
}

\lstdefinestyle{layout}{
	frame=none,
	captionpos=b,
}

\lstdefinestyle{comment-style}{
	morecomment=[l]//,
	morecomment=[s]{/*}{*/},
	commentstyle={\color{ccomment}\itshape},
}

\lstdefinestyle{string-style}{
	morestring=[b]",%
	morestring=[b]',%
	stringstyle={\color{cstring}},
	showstringspaces=false,%
}

\lstdefinestyle{keyword-style}{
	keywordstyle={\ttfamily\bfseries},
	morekeywords={
		function,
		constructor,
		int,
		bool,
		return,
		returns,
		uint
	},
	morekeywords = [2]{},
	keywordstyle = [2]{\text},
	sensitive=true,
}

\lstdefinestyle{input-encoding}{
	inputencoding=utf8,
	extendedchars=true,
	literate=
	{ℝ}{$\reals$}1%
	{→}{$\rightarrow$}1%
	{α}{$\alpha$}1%
	{β}{$\beta$}1%
	{λ}{$\lambda$}1%
	{θ}{$\theta$}1%
	{ϕ}{$\phi$}1%
}

\lstdefinestyle{escaping}{
	moredelim={**[is][\color{blue}]{\%}{\%}},
	escapechar=|,
	mathescape=true
}

\lstdefinestyle{default-style}{
	basicstyle=\fontencoding{T1}\ttfamily\footnotesize,
	style=numbers,
	style=layout,
	style=comment-style,
	style=string-style,
	style=keyword-style,
	style=input-encoding,
	style=escaping,
	tabsize=2,
	upquote=true
}

\lstdefinelanguage{BASIC}{
	language=C++,
	style=default-style
}[keywords,comments,strings]%

\usepackage[utf8]{inputenc} 
\usepackage[T1]{fontenc}    
\usepackage{caption}
\usepackage{enumerate}
\usepackage{url}            
\usepackage{booktabs}       
\usepackage{amsfonts}       
\usepackage{nicefrac}       
\usepackage{microtype}      
\usepackage{xcolor}         
\usepackage{tikz,booktabs,multirow,enumitem,bm}
\usepackage{colortbl}
\usepackage{tabularx}
\usepackage{threeparttable}
\usepackage{subcaption}
\usepackage{wrapfig}
\usepackage{tabulary}
\usepackage{amsmath,amsthm,amsfonts, bbm}
\usepackage{makecell}
\usepackage{rotating}
\usepackage{float}
\usepackage{array}
\usepackage{siunitx}
\usepackage[most]{tcolorbox}
\usepackage{CJKutf8}
\usepackage{pifont}
\newcommand{\cmark}{\ding{51}}
\newcommand{\xmark}{\ding{55}}

\usepackage{amsmath,amsfonts,bm}
\usepackage{amsthm}
\usepackage{amssymb}
\usepackage{mathtools}

\def\1{\bm{1}}

\def\vs{{\bm{s}}}

\DeclareMathAlphabet{\mathsfit}{\encodingdefault}{\sfdefault}{m}{sl}
\SetMathAlphabet{\mathsfit}{bold}{\encodingdefault}{\sfdefault}{bx}{n}

\newcolumntype{x}[2]{S[table-format=#1.#2,table-auto-round]}
\newcolumntype{y}[2]{>{\small} S[table-format=#1.#2,table-auto-round]}

\definecolor{hyperlinkblue}{HTML}{0000AA}
\hypersetup{citecolor=hyperlinkblue} 

\definecolor{red}{HTML}{FF0000}
\definecolor{drop}{HTML}{2596be}
\definecolor{anon}{HTML}{10a37f}
\definecolor{devil}{HTML}{84cf9d}

\lstdefinestyle{mystyle}{
    escapechar=\#,
    breaklines=true,
    basicstyle=\scriptsize\ttfamily,
    numbers=none,
    language={},
    framextopmargin=0pt,
    framexbottommargin=0pt,
    breakindent=0pt,
    showspaces = false,
    keywordstyle=\bfseries,
    showstringspaces=false,
    columns=fullflexible,
    morekeywords={Style, Consistency, Accuracy, Ethics, Score}
}

\newtcblisting{prompt}[2][]{
    arc=3pt, outer arc=3pt,
    width=.9\linewidth,
    left=1mm,
    top=0mm,
    bottom=0mm,
    title=#2, 
    colback=gray!5!white,
    colframe=black!75!black,
    fonttitle=\bfseries,
    listing only, 
    listing options={style=mystyle},
    breakable,
    #1
}

\newtcblisting{response}[2][]{
    arc=3pt, outer arc=3pt,
    width=.9\linewidth,
    left=1mm,
    top=0mm,
    bottom=0mm,
    title=#2, 
    colback=devil!5!white,
    colframe=devil!50!black,
    fonttitle=\bfseries,
    listing only, 
    listing options={style=mystyle,deletekeywords={Style}},
    breakable,
    #1
}

\newtcblisting{inference}[2][]{
    arc=3pt, outer arc=3pt,
    width=.9\linewidth,
    left=1mm,
    top=0mm,
    bottom=0mm,
    title=#2, 
    colback=red!5!white,
    colframe=red,
    fonttitle=\bfseries,
    listing only, 
    listing options={style=mystyle},
    breakable,
    #1
}

\newtcblisting{anonymize}[2][]{
    arc=3pt, outer arc=3pt,
    width=.9\linewidth,
    left=1mm,
    top=0mm,
    bottom=0mm,
    title=#2, 
    colback=anon!5!white,
    colframe=anon,
    fonttitle=\bfseries,
    listing only, 
    listing options={style=mystyle},
    breakable,
    #1
}

\lstdefinelanguage{QA}{
  morekeywords={Prompt,Base,Watermarked},
  sensitive=false,
}

\lstdefinestyle{qaStyle}{
  language=QA,
  basicstyle=\normalfont\small,      
  keywordstyle=\bfseries\color{teal},
  stringstyle=\color{black},         
  numbers=none,                       
  showstringspaces=false,            
  frame=single,                      
  rulecolor=\color{gray!50},         
  backgroundcolor=\color{gray!10},   
  columns=fullflexible,              
  xleftmargin=1em, xrightmargin=1em, 
  aboveskip=1em, belowskip=1em       
}

\definecolor{outerbg}{HTML}{F6F6F6}      
\definecolor{outerframe}{HTML}{D0D0D0}   

\definecolor{modelAback}{HTML}{DD8452}   
\definecolor{modelAframe}{HTML}{DD8452}  

\definecolor{modelBback}{HTML}{4C72B0}   
\definecolor{modelBframe}{HTML}{4C72B0}  

\newtcolorbox{promptbox}[1][]{%
  breakable,
  colback=white,
  colframe=black!25,
  fonttitle=\bfseries\small,
  coltitle=black!80,
  title=User~Prompt~{#1},
  boxsep=5pt,
  top=4pt,bottom=4pt,
  arc=1mm,
}

\newtcolorbox{modelbox}[3][]{%
  breakable,
  colback=#2!75,
  colframe=#3!85,
  coltitle=#3!10!black,
  fonttitle=\bfseries\small,   
  title={#1},
  boxsep=5pt,
  top=4pt,bottom=4pt,
  arc=1mm,
}

\BeforeBeginEnvironment{promptbox}{\par\nolinenumbers}
\AfterEndEnvironment{promptbox}{\linenumbers}
\BeforeBeginEnvironment{modelbox}{\par\nolinenumbers}
\AfterEndEnvironment{modelbox}{\linenumbers}

\definecolor{promptbg}{HTML}{F7F7F7}
\definecolor{promptframe}{HTML}{CCCCCC}
\definecolor{varcolor}{HTML}{9D2235}
\definecolor{schemabg}{HTML}{F0F4F8}
\definecolor{calibg}{HTML}{FFFEF5}

\newtcolorbox{promptbox2}[1][]{%
  colback=promptbg, colframe=promptframe,
  fonttitle=\small\bfseries, title={#1},
  boxrule=0.5pt, arc=2pt, left=6pt, right=6pt, top=4pt, bottom=4pt,
  fontupper=\small
}

\newtcolorbox{schemabox}[1][]{%
  colback=schemabg, colframe=promptframe,
  fonttitle=\small\bfseries\ttfamily, title={#1},
  boxrule=0.4pt, arc=1.5pt, left=6pt, right=6pt, top=3pt, bottom=3pt,
  fontupper=\small\ttfamily
}

\newtcolorbox{calibrationbox}[1][]{%
  colback=calibg, colframe=promptframe,
  fonttitle=\small\bfseries, title={#1},
  boxrule=0.4pt, arc=1.5pt, left=6pt, right=6pt, top=4pt, bottom=4pt,
  fontupper=\small
}

\BeforeBeginEnvironment{promptbox2}{\leavevmode\par\nolinenumbers}
\AfterEndEnvironment{promptbox2}{\linenumbers}
\BeforeBeginEnvironment{schemabox}{\leavevmode\par\nolinenumbers}
\AfterEndEnvironment{schemabox}{\linenumbers}
\BeforeBeginEnvironment{calibrationbox}{\leavevmode\par\nolinenumbers}
\AfterEndEnvironment{calibrationbox}{\linenumbers}

\newcommand{\pvar}[1]{{\color{varcolor}\texttt{\$#1}}}

\renewcommand{\S}{Sec.~}

\usepackage[capitalize,noabbrev]{cleveref}

\crefformat{section}{\S#2#1#3}

\crefrangeformat{section}{\S#3#1#4\crefrangeconjunction\S#5#2#6}

\crefmultiformat{section}{\S#2#1#3}{\crefpairconjunction\S#2#1#3}{\crefmiddleconjunction\S#2#1#3}{\creflastconjunction\S#2#1#3}

\newcommand{\crefrangeconjunction}{--}

\crefname{listing}{Lst.}{listings}
\crefname{line}{Lin.}{Lin.}
\crefname{appendix}{App.}{App.}

\newcommand{\appref}[1]{%
	\ifbool{includeappendix}{\cref{#1}}{the appendix}%
}
\newcommand{\Appref}[1]{%
	\ifbool{includeappendix}{\cref{#1}}{The appendix}%
}

\usepackage{aliascnt}
\theoremstyle{plain}

\newaliascnt{lemma}{theorem}

\aliascntresetthe{lemma}
\crefname{lemma}{Lemma}{Lemmas}
\Crefname{lemma}{Lemma}{Lemmas}

\theoremstyle{definition}

\theoremstyle{remark}

\newcommand{\OurTitle}{Marking the Wrong Symptoms: Evaluating LLM Watermarks in Medical Texts}

\title{\OurTitle{}}

\author{%
  \end{tabular}%
  \begin{minipage}{0.95\textwidth}\centering
    \textbf{Melanie Rieff\textsuperscript{1,$\ast$} \quad
    Robin Staab\textsuperscript{1} \quad
    Thibaud Gloaguen\textsuperscript{1}}\\[0.3em]
    \textbf{Stefan Hegselmann\textsuperscript{2,3} \quad
    Martin Vechev\textsuperscript{1}}\\[0.6em]
    \normalfont\normalsize
    \textsuperscript{1}Department of Computer Science, ETH Zurich, Switzerland \\
    \textsuperscript{2}Berlin Institute of Health at Charité -- Universitätsmedizin Berlin, Center of Digital Health, Berlin, Germany \\
    \textsuperscript{3}Deutsches Herzzentrum der Charité -- Medical Heart Center of Charité and German Heart Institute Berlin, Berlin, Germany \\[0.3em]
    \textsuperscript{$\ast$}Corresponding author: \texttt{mrieff@student.ethz.ch}
  \end{minipage}%
  \begin{tabular}[t]{@{}c@{}}%
}

\begin{document}

\maketitle


\begin{abstract}
	Large language models (LLMs) are increasingly integrated into clinical workflows, stressing the need for reliable traceability of model-generated output with watermarking. Yet, most watermarks are evaluated on general-purpose benchmarks, leaving domains like medicine, where small token-level perturbations can result in significant semantic changes, under-explored.
In this work, we present the first rigorous study of how LLM watermarks affect medical performance, benchmarking 5 watermarking schemes across 11 LLMs and 7 VLMs on various tasks spanning unimodal and multimodal clinical reasoning. Importantly, we complement existing evaluations by introducing a human-expert-validated pipeline for systematically auditing medical reasoning quality, terminological precision, and induced hallucinations.
Our results reveal that watermarking can induce substantial degradation across multiple failure modes, including lexical corruption, hallucinated terminology, and amplified misattribution or omission of image findings. Notably, we find that the absence of domain-specific analyses, combined with aggregate metrics that miss failures inherent to clinical text, can systematically obscure practical watermark-induced degradations. Our findings establish domain-specific evaluation as a prerequisite for the safe deployment of watermarked models in medicine, where current benchmarks can otherwise mask clinically consequential failures. 
\end{abstract}

\section{Introduction}
\label{sec:introduction}

Large language models (LLMs) are increasingly integrated into clinical workflows: 81\% of U.S.\ physicians now report using AI professionally~\citep{ama2026survey}, and LLM-based tools are actively explored for report generation, clinical decision support, and patient communication~\citep{gowda2025radllm, weissman2025unregulated}.
In particular, LLM outputs inform downstream judgement, from summarising patient histories and drafting reports to suggesting differential diagnoses.
As model capabilities improve, the scope of clinical reliance is likely to grow, making the quality of generated text a matter of patient safety.
In parallel, emerging~\citep{euaiact2024} regulation is pushing toward provenance marking of AI-generated content, with watermarking becoming the leading compliance mechanism for tracing LLM outputs~\citep{eucommission2026code}.
Crucially, this implies that LLMs used in clinical settings are increasingly likely to be using a watermark.

\paragraph{LLM Watermarking}
Watermarking embeds a statistically detectable signal by modifying the LLM sampling procedure at every generation step, which enables traceability but, by design, directly influences the model's output distribution.
Importantly, prior works evaluate the impact of watermarks on output quality with general-domain benchmarks, and coarse quality proxies such as perplexity.
In medicine, where correctness depends jointly on precise reasoning, terminology, qualifiers, and numerical values, even small token-level perturbations can alter clinical meaning significantly, suggesting that results from those simple evaluations might not transfer~\citep{tu2024waterbench, moll2025evaluating, hager2024evaluation}.
This makes understanding how watermarks alter LLMs' clinical performance a critical, yet under-explored issue.

\begin{figure}
    \centering
    \includegraphics[width=\linewidth]{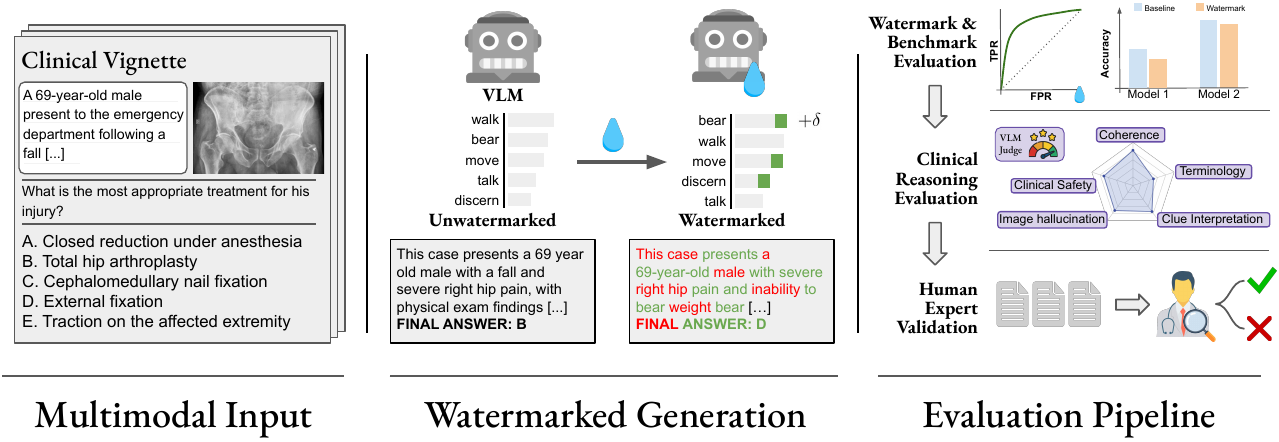}
    \caption{
        \textbf{Overview of our evaluation pipeline.}
        Given a multimodal benchmark instance, we prompt VLMs (using chain-of-thought prompting) with and without watermarks for reference. 
        We first evaluate the detectability of watermarks in the answers and their accuracy.
        We then examine the reasoning traces in more depth using three specific LLM-as-judges, validated by clinical experts.
        }
    \label{fig:hook}
\end{figure}

\paragraph{This work}
In this work, we propose the first systematic study of how text watermarking affects medical LLM performance, illustrated in~\cref{fig:hook}.
In particular, we evaluate $5$ prominent watermarks on two medical benchmarks, mixing clinical knowledge and visual understanding~(\cref{fig:hook} left). 
To measure the impact of watermarking on reasoning, we introduce~(in \cref{sec:evaluation_dimensions}) three specific LLM-as-judges validated against clinical expert annotations (\cref{fig:hook} right). 

Our results reveal that watermarking can leave benchmark accuracy largely intact while substantially degrading the underlying reasoning: the rate of correct answers backed by flawed reasoning more than doubles on several models (e.g.\ 11.4\%$\to$26.3\% on \textsc{Phi-4-14B} with distortionary SynthID), and fabricated medical entities increase by up to +39.2 pp.
Among questions where both watermarked and unwatermarked outputs select the correct answer, up to 40\% rely on mutually exclusive diagnostic statements.
On multimodal inputs, watermarks increasingly corrupt visual interpretation, and for reasoning models, watermarking the reasoning trace inflates output length by up to $69\%$ while degrading reasoning quality, whereas restricting the watermark to the final answer preserves both.

\paragraph{Key Contributions} Our main contributions are:
\begin{itemize}[leftmargin=1.4em]
      \item A systematic evaluation of five watermarking schemes across 11~LLMs and 7~VLMs on medical benchmarks (MedQA, MedXpertQA-MM), with \emph{physician-validated} LLM-as-a-judge audits that assess reasoning quality, terminological precision, and hallucinations beyond letter accuracy.
      \item The first analysis of how text-decoding watermarks interact with multimodal medical input, revealing that watermarking during text generation degrades visual grounding, \eg suppressing correct image-based claims and amplifying contradicted ones, despite accuracy remaining stable.
      \item A study of reasoning model watermarking, showing that restricting watermarks to the user-visible final answer preserves reasoning quality, while watermarking the thinking degrades terminological precision and induces circular deliberation.
\end{itemize}

\section{Related Work}
\label{sec:related_work}

We review current evaluation practices for LLMs in medicine, the main families of text watermarking, and existing work on watermark quality assessment, highlighting the gap that motivates our study.

\paragraph{LLMs in medicine.}
General-purpose and domain-specialized LLMs are commonly evaluated on USMLE-style benchmarks such as MedQA \citep{jin2021medqa, kim2025metareview}, while multimodal benchmarks \citep{hu2024omnimedvqa, liu2024gmaimmbench, zuo2025medxpertqa} extend evaluation to joint interpretation of images, labs, and patient records.
Crucially, benchmark accuracy alone is an insufficient measure of clinical quality: models may arrive at the correct answer through flawed reasoning \citep{maharana2025right}, produce confident hallucinations \citep{pal2023medhalt}, or misuse medical terminology while maintaining a fluent narrative.
These observations have motivated richer evaluation frameworks, from structured reasoning rubrics \citep{liu2025clever} to physician-graded benchmarks \citep{arora2025healthbench}, and underscore that even small prompt variations can substantially change medical output quality \citep{hager2024evaluation}.
Our work builds on this line by introducing 3 LLM-as-judges, validated against two board-certified physicians, specifically designed to audit watermark-induced degradation along the quality axes that matter in clinical~practice.

\paragraph{Watermarking.}
Generation-time text watermarking modifies the sampling procedure to embed statistically detectable signals tied to a secret key, enabling traceability of AI-generated content \citep{liu2024survey}.
At each token position, a context-dependent hash seeds a pseudorandom score assignment over the vocabulary, and the sampling distribution is modified accordingly.
\emph{Distortionary} methods such as KGW \citep{kgw1} and PPL \citep{gloaguen2026unifiedframeworkllmwatermarks} bias next-token distributions toward high-scoring tokens, introducing a systematic shift away from the model's original distribution.
\emph{Distortion-free} methods such as DipMark \citep{dipmark}, AAR at zero strength \citep{aar}, and SynthID-Text with two leaves per tournament \citep{synthid} instead resample tokens in a way that preserves the original model distribution in expectation over the random watermark key.
However, this guarantee is an \emph{expectation} over keys, and under the assumption of no hash collisions.
We find that, in practice, even distortion-free schemes can induce some small reasoning degradation~(\cref{sec:reasoning_degradation}).

\paragraph{Evaluating watermark quality.}
Watermarking studies typically report detection power alongside coarse quality proxies such as perplexity or brief LLM ratings \citep{kgw1, synthid, kth}. 
Yet, these proxies are poorly aligned with human preference \citep{tu2024waterbench} and mask substantial, task-dependent degradation: instruction-following quality drops considerably under matched watermark strength \citep{tu2024waterbench}, smaller models suffer disproportionately \citep{yang2025contextual}, and reasoning-intensive tasks degrade more severely than open-ended generation \citep{lee2024sweet, chen2024watme}.
These findings establish that the impact of watermarking is model- and task-specific, yet most evaluations are confined to general-domain benchmarks, leaving a twofold gap: (i)~the tasks studied do not capture the demands of clinical text, where a single-token substitution can alter a diagnosis, and (ii)~the quality proxies employed cannot detect failure modes such as terminology corruption, hallucinated findings, or inconsistency between reasoning and conclusion.
Separately, recent work has shown that watermarks can be applied to reasoning traces \citep{PrincipleSemantic2026Reasoning, chen2024watme}, yet no study has assessed whether such application preserves clinical reasoning quality. 
One prior study attempts to evaluate watermarking on medical text~\citep{hastuti2025factuality}.
Yet, we find that it suffers from fundamental methodological flaws, which makes its findings unreliable.
Namely, it uses instruction-tuned models without chat templates, caps generation at 25 tokens, too short for reliable watermark detection, and fails to establish a paired unwatermarked baseline from which to measure hallucinations.

Our work introduces, for the first time, a rigorous and systematic evaluation of watermarks in medicine, spanning five watermarking schemes, multiple model scales and domain specializations, and text-decoding watermarks applied to models that process multimodal medical inputs.
Importantly, we go beyond accuracy and introduce 3 LLM-as-judges, validated by clinical experts, to probe reasoning quality, lexical precision, and diagnostic consistency.

\section{Our Evaluation Pipeline}
\label{sec:experiments}

In this section, we introduce our evaluation pipeline illustrated in \cref{fig:hook}.
We describe the models, watermarks, and data we evaluate~(\cref{sec:experimental_setup}), and the metrics we use~(\cref{sec:evaluation_dimensions}).
We defer additional details on the clinician validation of our LLM judges to~\cref{app:clinician_validation}.

\subsection{Setup}
\label{sec:experimental_setup}

Our evaluation spans two medical benchmarks probing complementary aspects of clinical knowledge, 18 language and vision--language models covering general-purpose and medical-specialised architectures at 4B--70B scale, and five prominent watermarking schemes.
Together these define the (model, watermark, strength) configurations we evaluate throughout the next sections.

\paragraph{Benchmarks.}
We anchor our evaluation with two medical benchmarks, each of them chosen to probe distinct aspects of medical knowledge and reasoning.

\emph{MedQA}~\citep{jin2021medqa} contains 2{,}000 four-option, multiple-choice questions drawn from the United States medical licensing examinations.
A meta-analysis by \citet{kim2025metareview} identifies MedQA as the benchmark most predictive of real-world clinical performance among existing medical QA datasets.

\emph{MedXpertQA-MM}~\citep{zuo2025medxpertqa} comprises 2{,}000 expert-curated, five-option questions spanning 17 medical specialties. Each question is paired with one to six images
(on average 1.43, across 10 modalities including radiology, pathology, and schematic diagrams), requiring joint interpretation
alongside clinical vignettes and structured patient data such as laboratory results.

\paragraph{Models.}
We evaluate three groups of models:
On \emph{MedQA}, we include six instruction-tuned LLMs: 4 general-purpose models (\textsc{Llama-3.1-8B}, \textsc{Llama-3.1-70B}~\citep{grattafiori2024llama3}, \textsc{Gemma-3-12B}~\citep{team2025gemma3}, and \textsc{Phi-4-14B}~\citep{abdin2024phi4}), and 2 medical-specialised models (\textsc{OpenBioLLM-70B}~\citep{pal2024openiollms}, and \textsc{UltraMedical-70B}~\citep{zhang2024ultramedical}), as well as 4 reasoning models (\textsc{DeepSeek-R1-Distill-Llama-8B}, \textsc{DeepSeek-R1-Distill-Llama-70B}, \textsc{DeepSeek-R1-Distill-Qwen-32B}~\citep{guo2025deepseek}, and \textsc{Qwen-3.5-27B}~\citep{qwen2025qwen35}).
On \emph{MedXpertQA-MM}, we evaluate 7 VLMs spanning 4B--32B parameters: \textsc{Qwen-3-VL-8B} and \textsc{Qwen-3-VL-32B}~\citep{qwen2025qwen3}, \textsc{Gemma-4-31B}~\citep{team2026gemma4}, \textsc{Lingshu-7B} and \textsc{Lingshu-32B}~\citep{xu2025lingshu}, and \textsc{MedGemma-4B} and \textsc{MedGemma-27B}~\citep{sellergren2025medgemma}.
In particular, \textsc{Lingshu-7B}, \textsc{Lingshu-32B}, \textsc{MedGemma-4B}, and \textsc{MedGemma-27B} are medical-specialised VLMs and were not trained on MedXpertQA.

\paragraph{Watermarking schemes.}
Five prominent generation-time watermarking schemes are evaluated, spanning the major algorithmic families: KGW~\citep{kgw1} (soft logit bias), DipMark~\citep{dipmark} (CDF reweighting), AAR~\citep{aar} (Gumbel-max sampling), PPL~\citep{gloaguen2026unifiedframeworkllmwatermarks} (perplexity-budgeted argmax), and SynthID~\citep{synthid} (tournament sampling).
All five share a common structure: at each token position, a context-dependent hash assigns pseudorandom scores to vocabulary entries, and the sampling distribution is modified accordingly.
The schemes fall into two categories.
\emph{Distortion-based} methods (KGW, PPL at high strength) bias next-token distributions toward high-scoring tokens, introducing a systematic shift away from the model's original distribution.
\emph{Distortion-free} methods (DipMark, AAR at zero strength, SynthID with binary tournaments) preserve the original distribution in expectation over the random watermark key.
For each scheme, we sweep multiple strength levels to evaluate their detectability--quality trade-off. 
We describe each scheme in-depth in \cref{app:watermark-details}.

\subsection{Evaluation Dimensions}
\label{sec:evaluation_dimensions}

Each (model, watermark, strength) configuration is scored along two complementary fronts: (1) baseline metrics that capture the watermark strength and benchmark utility, and (2) structured audits that probe whether watermarking corrupts underlying medical reasoning.
The audits are necessary as benchmark accuracy fails to detect fabricated entities, misapplied terminology, or distorted reasoning.

\paragraph{Baseline metrics.}
We sample each (model, watermark, strength) configuration on the full 2{,}000-question benchmark using chain-of-thought prompting (with prompts in \cref{app:prompts}).
For each completion, we measure both the \emph{accuracy} and the watermark \emph{detectability}.
For watermark detectability, we measure the average true positive rate (TPR) at a $1\%$ false positive rate (FPR) over the 2{,}000 answers.

\paragraph{Reasoning-quality LLM-Judge.}
A model may select the correct answer while relying on fabricated entities or misapplied terminology, a failure mode documented in prior work on medical LLMs~\citep{maharana2025right,pal2023medhalt} and one that standard accuracy metrics cannot detect.
Letter accuracy alone therefore cannot show whether watermarking degrades medical reasoning.
We complement accuracy with three structured LLM-as-judges, using \textsc{Gemini-3-Flash} as the judge for its balance of cost and reasoning capability.
\begin{enumerate}
    \item \emph{Single-response audit} evaluates each completion in isolation: the judge sees one response at a time, blind to both watermark condition and gold answer, and flags eight axes (fabricated entities, corrupted spellings, and misapplied real terms (terminology), confident error, vignette fabrication, clue misread, within-response contradiction, linguistic corruption, and format/termination failure) plus an abstention indicator.
    For MedXpertQA-MM, the judge additionally classifies each perceptual image-grounding claim as \textsc{supported}, \textsc{contradicted}, or \textsc{unverifiable}.
    \item \emph{Pairwise divergence audit} presents the watermarked and unwatermarked completions for the same question side by side to the judge, with response order randomised and condition labels stripped.
    The judge records the correctness configuration of the pair, the level of diagnostic interpretation divergence (\textsc{none}, \textsc{minor}, \textsc{major}), and any scenario fabrication.
    \item \emph{Reasoning-trace audit} targets thinking models, whose hidden chain-of-thought may be independently distorted by the watermark. We use a two-block judging protocol. First, without seeing the gold answer, the judge rates four aspects of reasoning quality: reasoning efficiency, clinical coherence, alignment between the reasoning and the final answer, and engagement with distractors. Second, with the gold answer, the judge rates two factual aspects: terminological precision and diagnostic accuracy. The judge also provides two overall clinical-deployment ratings, \emph{clinical safety} and \emph{supervisory burden}, as well as the primary degradation pattern, defined in \cref{app:judge_prompts}.
    To separate degradation in the reasoning trace from degradation in the final answer, we run two paired comparisons: (i) \emph{full watermark vs.\ final-answer-only watermark} and (ii) \emph{final-answer-only watermark vs.\ no watermark}.
\end{enumerate}

Importantly, we validate our judges with physician agreement in \cref{sec:human_validation,app:judge_validation}.
For each configuration, we conduct judge audits at the watermark hyperparameters that yield the lowest strength while still achieving $\geq 99\%$ TPR at $1\%$ FPR, which prior work has considered the operating point required for regulatory deployment.
We defer the remaining judge parameters (\eg full prompts, axis definitions, calibration examples, and decoding configurations) to \cref{app:judge_prompts}.

\section{Results}
\label{sec:results}

In this section, we evaluate how prominent watermarks affect the clinical performance of LLMs and validate our judges.
We show that even though pure benchmark accuracy may remain high~(\cref{sec:accuracy}), watermarks can still degrade reasoning quality~(\cref{sec:reasoning_degradation}).
We then validate our judges in \cref{sec:human_validation}.
Lastly, in \cref{sec:thinking}, we evaluate watermarks in the context of reasoning models.

\subsection{Accuracy Looks Mostly Stable, But the Picture is Misleading}
\label{sec:accuracy}


\begin{figure}[t]
    \centering
    \includegraphics[width=\textwidth]{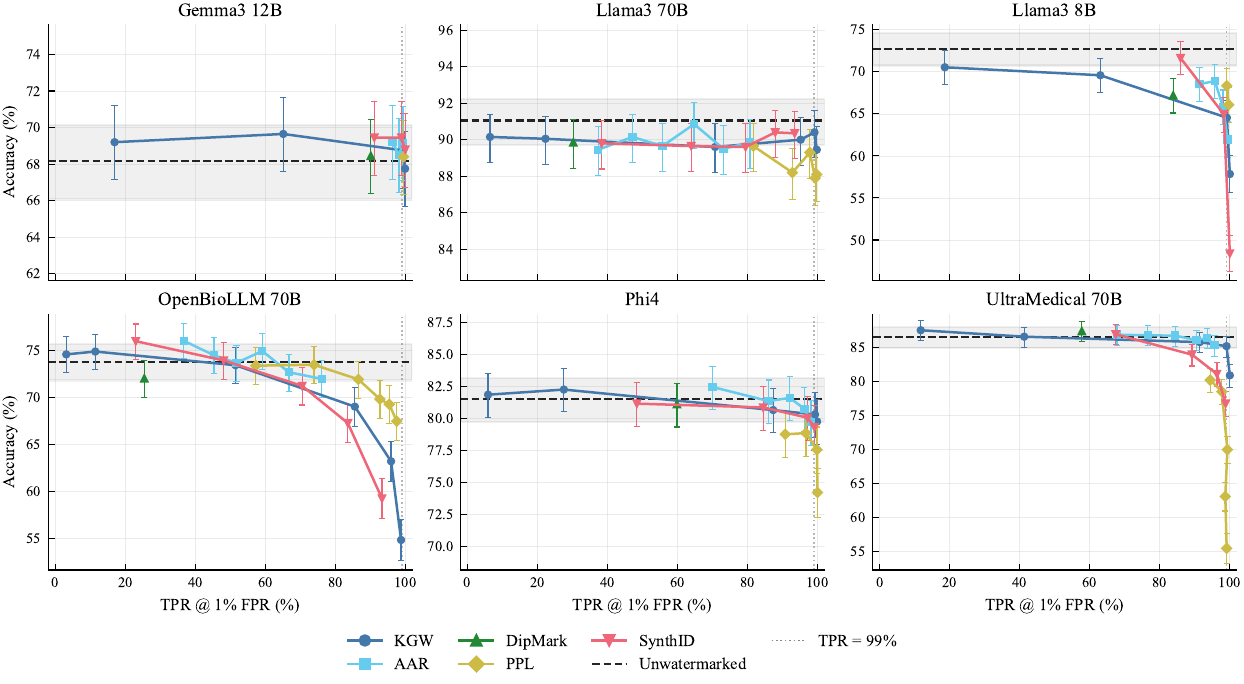}
    \caption{MedQA accuracy vs.\ detectability (TPR@$1\%$ FPR), non-reasoning LLMs across five schemes. Dashed line and grey band: unwatermarked mean and $95\%$ CI; dotted vertical: TPR\,$=99\%$. \textsc{Llama-3.1-70B}, \textsc{Gemma-3-12B}, and \textsc{Phi-4-14B} stay within the band across all schemes; the two specialised 70Bs collapse under complementary schemes; \textsc{Llama-3.1-8B} degrades broadly.}
    \label{fig:tpr_acc_medqa}
\end{figure}

\begin{figure}[t]
    \centering
    \includegraphics[width=\textwidth]{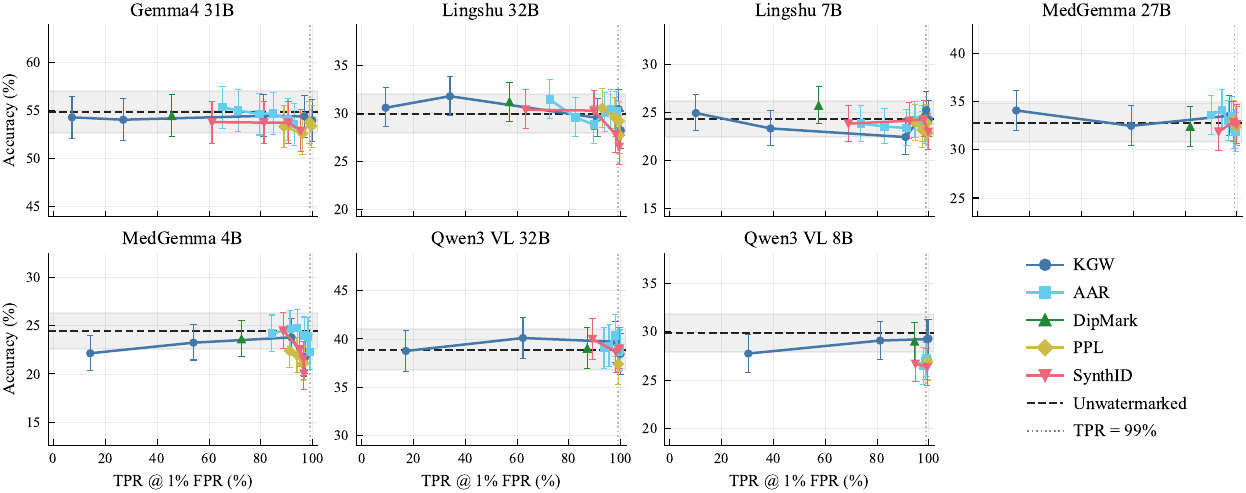}
    \caption{MedXpertQA-MM accuracy vs.\ detectability, seven VLMs; conventions as in \cref{fig:tpr_acc_medqa}. Models consistently stay near the baseline band despite reasoning varying noticeably (\cref{sec:reasoning_degradation}).}
    \label{fig:tpr_acc_medxpert}
\end{figure}

\cref{fig:tpr_acc_medqa,fig:tpr_acc_medxpert} show that, for most schemes and models, accuracy remains within the 95\% confidence interval (CI) of the unwatermarked baseline.
On MedQA~(\cref{fig:tpr_acc_medqa}), three configurations show a visible drop: the medical-specialised 70Bs (\textsc{UltraMedical-70B} and \textsc{OpenBioLLM-70B}) at high TPR@1\%FPR, and \emph{\textsc{Llama-3.1-8B}} more broadly.
That two equally large medical-specialised models break at high TPR@1\%FPR, while the equally large general-purpose \textsc{Llama-3.1-70B} remains consistently stable, rules out scale or domain specialisation as sufficient explanations.\footnote{\emph{\textsc{BioMistral-7B}} also collapses (KGW $-37.5$, SynthID $-14.2$\,pp), but the signal is confounded by poor instruction tuning on the CoT-MCQ template (median $\sim 6$ words per response vs.\ $250$--$450$ for other models); we exclude it from the headline claims and report it only in the appendix grids.}
On MedXpertQA-MM~(\cref{fig:tpr_acc_medxpert}), all VLMs stay within or near the CI.
We find in~\cref{app:ppl} that although perplexity rises monotonically with watermark strength on every model, it does not correlate with drops in accuracy or reasoning degradation.
Hence, a standard accuracy benchmark would conclude that watermarking is safe for medical use.
Our audits in \cref{sec:reasoning_degradation} show that this conclusion is incorrect: even where accuracy holds, watermarking corrupts the reasoning that clinicians would actually read.

\subsection{Reasoning Quality Degrades Even When Accuracy Holds}
\label{sec:reasoning_degradation}

We score every (model, scheme) configuration at the $\geq 99\%$ TPR operating point using the judge audits of \cref{sec:evaluation_dimensions}.
We bootstrap all our results, and our p-values quantify the significance between watermarked and unwatermarked results.
We show the two most-affected models per benchmark in \cref{tab:main_combined}, and defer full results to \cref{app:full_judge_tables}.

\paragraph{Correct answers hide broken reasoning.}
Watermarking increases the rate at which models reach the right MCQ letter through flawed reasoning.
The \emph{faulty-but-correct} rate (F\&C) (\ie correct letter with at least one flagged defect) climbs from $11.4\%$ to $26.3\%$ on \textsc{Phi-4-14B} under SynthID while accuracy drops only $-3.2$\,pp, and similarly on \textsc{OpenBioLLM-70B} F\&C more than doubles ($12.9\% \to 27.5\%$, ${}^{*}$).
However, F\&C only points out the existence of inconsistencies within the reasoning, but does not explain exactly what the inconsistencies are.

\paragraph{Watermarking silently corrupts medical language.}
Even when accuracy and F\&C barely move, watermarks inject three categories of medical-language errors that standard metrics miss: \emph{fabricated entities}, \emph{misapplied real terms}, and \emph{corrupted spellings} (\cref{tab:terminology_examples}).
On \textsc{MedGemma-4B} under SynthID, $\Delta$F\&C is only $+0.4$, yet the model fabricates ${+12.6}$ extra entities per $100$ questions and misapplies ${+24.3}$ real terms; on \textsc{Lingshu-7B} under SynthID, fabrications reach ${+39.2}$ per $100$ questions.
Fabricated and misapplied terms are especially dangerous: they produce fluent, confident text that requires domain expertise to catch and that no spell-checker or final-letter check would flag.


\begin{table}[t]
\centering
\caption{\textbf{Illustrative watermark-induced terminology errors} drawn from our judge audits and paired qualitative reviews of watermarked outputs. Matched unwatermarked completions on the same questions use the correct terms.}
\label{tab:terminology_examples}
\scriptsize
\setlength{\tabcolsep}{4pt}
\renewcommand{\arraystretch}{1.15}
\begin{tabularx}{\textwidth}{@{} >{\raggedright\arraybackslash}p{1.2cm} >{\raggedright\arraybackslash}p{4.2cm} X @{}}
\toprule
\textbf{Error type} & \textbf{Watermarked Term} & \textbf{Correct Term / Explanation} \\
\midrule
\multirow{4}{1.2cm}{Fabricated entity}
  & \emph{Streptococcus epidermidis}
  & \emph{Staphylococcus epidermidis} \\
  & Pentaprazole
  & Pantoprazole \\
  & Naftifloxacin
  & No such antibiotic. \\
  & plutonium channels
  & potassium channels; plutonium is a radioactive element. \\
\midrule
\multirow{3}{1.2cm}{Misapplied real term}
  & Eosinophilic granuloma applied to the brown tumors of hyperparathyroidism
  & Eosinophilic granuloma denotes Langerhans cell histiocytosis---unrelated to the lytic lesions of hyperparathyroidism. \\
  & ``$\beta$-blockers preferred over CCBs in COPD''
  & Reversed: $\beta$-blockers risk bronchospasm; non-dihydropyridine CCBs are safe in COPD. \\
  & Benzodiazepines ``increase the \emph{duration} of GABA-channel opening''
  & Benzodiazepines increase the channel-opening \emph{frequency}; \emph{duration} is the mechanism of barbiturates. \\
\midrule
\multirow{3}{1.2cm}{Corrupted spelling}
  & \emph{Clostridioides diffidile}
  & Misspelling of \emph{C.\ difficile}. \\
  & meningritis
  & Misspelling of meningitis. \\
  & bradiceardia
  & Misspelling of bradycardia. \\
\bottomrule
\end{tabularx}
\end{table}

\begin{table}[t]
\centering
\caption{\textbf{Reasoning damage at $99\%$ TPR}, for two most-affected models per benchmark with full results in \cref{app:full_judge_tables}. 
We report the accuracy ($\Delta$Acc), the percentage of fabricated facts (Fab/100), the percentage of misapplied medical terms (Misapp/100), the percentage of corrupted spellings (Spell/100), the vignette fabrication ratio (VigFab), the faulty-but-correct rate (F\&C), and the percentage of paired responses with major diagnostic divergence (Major\%).
For the unwatermarked row we report the actual value, whereas for watermarked rows we report the difference/ratio with the unwatermarked baseline.
We \textbf{bold} entries that are statistically significant (p-value below $0.01$).
}
\label{tab:main_combined}
\footnotesize
\setlength{\tabcolsep}{3pt}
\begin{threeparttable}
\begin{tabular}{llrrrrrrr}
\toprule
Model & Scheme & $\Delta$Acc & Fab/100 & Misapp/100 & Spell/100 & VigFab & F\&C & Major\% \\
\midrule
\multicolumn{9}{l}{\textit{MedQA}} \\
\rowcolor{black!8} \textsc{Phi-4-14B} & Unwatermarked & 81.4 & 0.3 & 4.1 & 0.1 & 1.0$\times$ & 11.4 & -- \\
 & AAR & -2.5 & \textbf{+1.8} & \textbf{+4.0} & \textbf{+1.5} & \textbf{14.0$\times$} & \textbf{+6.5} & 1.9 \\
 & DipMark & -0.9 & +0.1 & +2.6 & +0.3 & 0.8$\times$ & \textbf{+3.6} & 2.8 \\
 & KGW & -0.7 & \textbf{+1.5} & \textbf{+3.6} & \textbf{+2.2} & \textbf{3.0$\times$} & \textbf{+9.5} & 3.6 \\
 & PPL & \textbf{-8.6} & \textbf{+3.8} & +4.3 & \textbf{+69.6} & \textbf{18.0$\times$} & \textbf{+9.6} & 2.3 \\
 & SynthID & \textbf{-3.2} & \textbf{+6.6} & \textbf{+8.0} & \textbf{+10.6} & \textbf{7.6$\times$} & \textbf{+14.9} & 3.3 \\
\rowcolor{black!8} \textsc{OpenBioLLM-70B} & Unwatermarked & 73.6 & 0.0 & 4.5 & 0.6 & 1.0$\times$ & 12.9 & -- \\
 & AAR & -0.2 & \textbf{+0.9} & +1.5 & +0.9 & 0.9$\times$ & \textbf{+4.4} & 4.4 \\
 & DipMark & -1.1 & +0.4 & +0.8 & +0.3 & 1.1$\times$ & +1.5 & 3.2 \\
 & KGW & \textbf{-17.4} & \textbf{+1.9} & \textbf{+4.4} & \textbf{+5.2} & \textbf{3.5$\times$} & \textbf{+4.2} & 5.6 \\
 & PPL & \textbf{-7.7} & \textbf{+2.1} & +1.8 & \textbf{+51.5} & 1.9$\times$ & \textbf{+6.7} & 6.6 \\
 & SynthID & \textbf{-11.4} & \textbf{+13.6} & \textbf{+6.9} & \textbf{+25.9} & \textbf{6.6$\times$} & \textbf{+14.6} & 8.7 \\
\multicolumn{9}{l}{\textit{MedXpertQA-MM}} \\
\rowcolor{black!8} \textsc{MedGemma-4B} & Unwatermarked & 24.1 & 1.9 & 11.1 & 1.0 & 1.0$\times$ & 16.5 & -- \\
 & AAR & -1.0 & \textbf{+6.5} & +8.1 & \textbf{+3.4} & \textbf{3.1$\times$} & +1.2 & 38.3 \\
 & DipMark & +0.2 & +0.7 & +2.0 & +0.2 & 1.2$\times$ & +1.6 & 35.0 \\
 & KGW & -2.6 & \textbf{+8.9} & +16.2 & \textbf{+15.4} & \textbf{3.1$\times$} & +0.6 & 31.7 \\
 & PPL & -1.7 & \textbf{+6.0} & +5.0 & \textbf{+43.6} & \textbf{2.1$\times$} & 0.0 & 34.7 \\
 & SynthID & -3.1 & \textbf{+12.6} & \textbf{+24.3} & \textbf{+18.0} & \textbf{4.2$\times$} & +0.4 & 36.0 \\
\rowcolor{black!8} \textsc{Lingshu-7B} & Unwatermarked & 23.9 & 1.5 & 12.1 & 0.4 & 1.0$\times$ & 13.7 & -- \\
 & AAR & -0.5 & \textbf{+12.9} & +18.1 & \textbf{+5.4} & \textbf{2.8$\times$} & +3.7 & 33.0 \\
 & DipMark & +1.0 & +0.3 & +2.7 & +0.1 & 1.0$\times$ & +1.4 & 22.6 \\
 & KGW & -0.8 & \textbf{+13.2} & +2.4 & \textbf{+16.2} & \textbf{3.6$\times$} & +6.0 & 42.9 \\
 & PPL & -0.6 & \textbf{+3.9} & +11.5 & \textbf{+37.9} & \textbf{2.4$\times$} & +2.3 & 20.0 \\
 & SynthID & -2.1 & \textbf{+39.2} & \textbf{+20.2} & \textbf{+18.9} & \textbf{7.0$\times$} & \textbf{+5.5} & 30.5 \\
\bottomrule
\end{tabular}
\end{threeparttable}
\vspace{-0.15in}
\end{table}

\paragraph{Each scheme damages a different axis.}
We find in \cref{tab:taskA_full_MedQA} that watermarking is not generic noise: each scheme damages a different axis.
PPL drives surface corruption.
SynthID and KGW instead inflate fabrications and term misuse, with vignette-fabrication ratios up to $18\times$ (\textsc{Phi-4-14B}/PPL) and $7.6\times$ (\textsc{Phi-4-14B}/SynthID).
AAR produces broad small-to-moderate effects.
DipMark causes the least damage, consistent with its distortion-free guarantee~\citep{dipmark}.
Yet distortion-free does not mean benign: the guarantee holds in expectation over the random key and without hash collisions.
We find that, at a fixed key, \textsc{Phi-4-14B} under DipMark still shows a statistically significant increase in F\&C.

\begin{wrapfigure}{R}{0.5\textwidth}
    \vspace{-0.3\baselineskip}
    \centering
    \includegraphics[width=\linewidth]{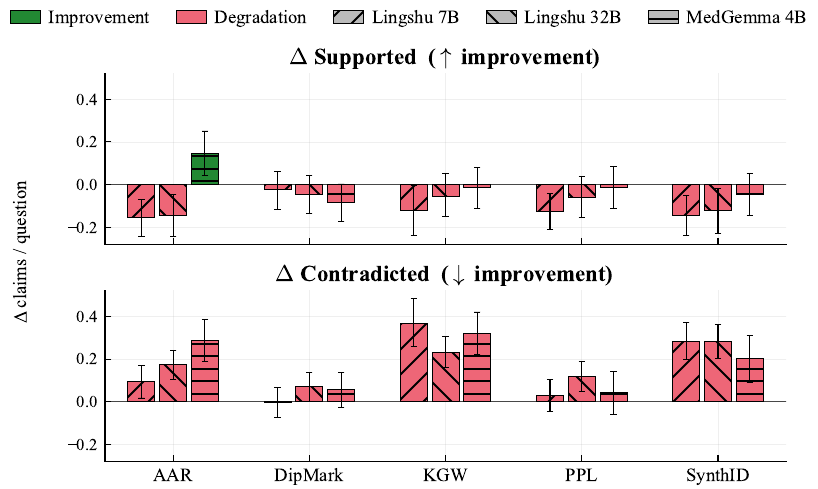}
    \caption{\textbf{Impact of watermarking on visual understanding.} We show the difference between watermarked and unwatermarked reasoning in the average number of supported (top) and contradicted (bottom) visual claims on MedXpertQA.}
    \label{fig:image_decomp}
    \vspace{-2\baselineskip}
\end{wrapfigure}

\paragraph{Watermarks corrupt visual reporting.}
Here, we find that watermark-induced degradation may change what models claim to see.
On MedXpertQA-MM, we track supported and contradicted visual claims per question against the provided image (\cref{fig:image_decomp}).
\textsc{supported} claims fall and \textsc{contradicted} claims rise simultaneously in most configurations with medical VLMs.
For the larger general-purpose VLMs, we find in \cref{app:dose_response} no significant change.

\paragraph{Domain-specialised and smaller models are most vulnerable.}
At matched scale and TPR@1\%FPR, general-purpose models are less degraded by watermarking than medical-specialised or small models.
\textsc{Llama-3.1-70B} is stable under all five schemes while the two same-size medical 70Bs each degrade on at least one scheme.
We find in \cref{app:full_judge_tables} a similar split among $30$B VLMs (\textsc{Gemma-4-31B} vs.\ \textsc{Lingshu-32B}).
Yet, at higher watermark distortion, even \textsc{Llama-3.1-70B}'s F\&C nearly doubles and \textsc{Gemma-4-31B}'s vignette fabrication reaches $6\times$.

\paragraph{Harm grows with strength, but the trajectory is scheme-specific.}
We find in~\cref{app:dose_response,fig:dose_response} that the quality-detectability trajectory (\ie how medical degradation varies with each scheme's detectability) is scheme-specific.
For instance, KGW degrades smoothly with detectability whereas SynthID's quality matches the unwatermarked baseline up to $90$\% TPR@1\%FPR and then degrades.

\paragraph{Understanding why watermarks degrade reasoning}
We observed that the failures above share a common underlying cause: when the watermark perturbs the sampling distribution at a token where the model commits to a clinical interpretation, it can redirect the entire downstream reasoning chain.
We isolate this with a controlled prefix-seeding experiment on three MedXpertQA-MM questions (\cref{app:branching}) and two schemes (KGW and PPL): watermarking amplifies distractor prefixes the clean model dismisses (accuracy drops of 14--25\,pp) and suppresses rescue cues it would otherwise exploit.

\subsection{Judge Validation}
\label{sec:human_validation}

A board-certified physician annotated $650$ completions blinded to model and watermark configuration, drawn from four (model, scheme) configurations spanning the full error-density range: \textsc{Phi-4-14B}/SynthID and \textsc{Llama-3.1-70B}/DipMark on MedQA, \textsc{Lingshu-7B}/SynthID and \textsc{Gemma-4-31B}/DipMark on MedXpertQA-MM ($250$ MedQA $+$ $250$ MedXpertQA-MM single-response completions, plus $150$ pairwise pairs).
As a second automated baseline, \textsc{Claude-Sonnet-4.6} scores the same $650$ samples independently.
 
On the four axes that carry our headline claims, judge--physician agreement is substantial: fabricated-entity counts reach $\kappa{=}0.75$ / $\rho{=}0.75$ ($n{=}500$), misapplied-term counts $\kappa{=}0.42$ / $\rho{=}0.47$, image-contradicted claims $\rho{=}0.74$ ($n{=}250$, MM only), and pairwise diagnostic divergence $\kappa_w{=}0.65$ / binary $\kappa{=}0.67$ ($n{=}150$).
Agreement between \textsc{Gemini-3-Flash} and \textsc{Claude-Sonnet-4.6} on the medical-content axes (fabrications, misapplied terms, pairwise divergence) matches or exceeds the corresponding judge--physician values ($0.80$ / $0.81$, $0.50$ / $0.52$, $\kappa_w{=}0.66$ / $\kappa{=}0.72$, respectively), confirming that two independently trained LLM-judges recover the same signal.
The full per-axis agreement grid (\cref{tab:judge_agreement_full}) shows that lower $\kappa$ values concentrate on low-prevalence stylistic flags (format/termination, linguistic corruption, spelling), where judge--physician and judge--judge agreement drop jointly---consistent with base-rate sensitivity of $\kappa$ rather than systematic bias.
We therefore anchor claims on these axes to paired $\Delta$s between watermarked and unwatermarked conditions rather than to absolute error levels.


\subsection{Reasoning Models: Full vs.\ Answer-Only Watermarking}
\label{sec:thinking}

The two highest-scoring models on MedQA in our setup are reasoning models (\textsc{Qwen-3.5-27B} ($94.6\%$) and \textsc{DeepSeek-R1-Distill-Llama-70B} ($92.3\%$)), narrowly above the strongest instruction-tuned baseline (\textsc{Llama-3-70B} ($91.0\%$)).
Unlike instruction-tuned models, they generate a reasoning trace (usually hidden) inside \texttt{<thinking>} tags before producing the final answer; a watermark can therefore be applied in two distinct ways: \emph{answer-only} (FA-only), which watermarks only the post-\texttt{<thinking>} answer tokens, and \emph{full watermark} (Full~WM), which watermarks both the reasoning trace and the final answer.
We compare both approaches on four reasoning models--\textsc{DeepSeek-R1-Distill-Llama-8B} and \textsc{DeepSeek-R1-Distill-Llama-70B}, \textsc{DeepSeek-R1-Distill-Qwen-32B}, and \textsc{Qwen-3.5-27B} with KGW at 3 strength levels ($\delta \in \{1, 3, 5\}$).

\paragraph{Full~WM degrades reasoning quality on every axis except the final answer.}
Across the six judge dimensions, Full~WM drives substantial negative shifts on \emph{reasoning efficiency} and \emph{terminological precision}, with smaller drops on \emph{clinical coherence} and \emph{distractor engagement} for each of the four models (\cref{fig:thinking_dim_scores}).
Two dimensions change negligibly: \emph{answer--reasoning alignment} and \emph{diagnostic accuracy}.
The model thus still selects the correct letter through reasoning that is internally consistent with its conclusion, but its output becomes substantially harder to use as a clinical reference due to imprecise or fabricated terminology, redundant deliberation, and distractor engagement that the unwatermarked response dismisses.
We also rate how harmful the reasoning is (\textsc{no\_harm\_risk}/\textsc{low\_risk}/\textsc{harm\_risk\_present}).
Even though both small and large models similarly degrade the reasoning, only small models' reasoning is judged to be significantly harmful.

\paragraph{Watermarking only the final answer is essentially free.}
Scoring Full~WM against FA-only as the reference reproduces the same per-dimension pattern within bootstrap intervals (\cref{fig:thinking_dim_scores}, orange vs.\ blue): FA-only is equivalent to Base on every axis.
\textsc{harm\_risk\_present} likewise stays at the No~WM baseline under FA-only on every model (\cref{tab:thinking_safety}): the watermark, applied only to post-\texttt{<thinking>} answer tokens, leaves the model's chain of thought unperturbed.  

\begin{wraptable}{r}{0.52\linewidth}
  \centering
  \footnotesize
  \vspace{-1\baselineskip}
  \begin{tabular}{@{}p{0.46\linewidth} p{0.46\linewidth}@{}}
    \toprule
    \textbf{Baseline (No~WM)} & \textbf{Full~WM (KGW $\delta=3$)} \\
    \midrule
    \emph{Linear:} ``The patient has low Vitamin D but normal
       Ca\textsuperscript{++}, suggesting secondary hyperparathyroidism.''
    & \emph{Circular:} ``Wait, let me rethink. The
       Ca\textsuperscript{++} is normal. Actually, the
       Ca\textsuperscript{++} is not low. Hmm. Let me check the
       Ca\textsuperscript{++} again.'' \\
    \bottomrule
  \end{tabular}
  \caption{Excerpt of \textsc{DeepSeek-R1-Distill-Qwen-32B} reasoning traces, baseline vs.\ Full~WM.}
  \label{tab:thinking_calcium}
  \vspace{-1\baselineskip}
\end{wraptable}

\paragraph{Trace watermarking inflates verbosity and induces recursive self-correction loops.}
Inspection of Full~WM traces reveals a consistent pathology across the four models: revisiting already established (and corrected) premises.
A representative excerpt is shown in \cref{tab:thinking_calcium}.
Quantitatively, Full~WM at $\delta = 5$ inflates mean total output length (reasoning trace plus final answer) by $+36$\,\% to $+69$\,\% across the four models, while FA-only does not~(\cref{tab:thinking_acc_tpr_length}).
This means that Full~WM's increase in detectability (at a given distortion level) is partly a length artefact rather than a stronger per-token signal.

\section{Conclusion and Limitations}
\label{sec:limitations}

We introduce a rigorous and systematic evaluation of watermark-induced degradation in medicine, with three specific LLM-as-judges validated by clinicians.
We find that, with most watermarks and models tested, the accuracy on medical benchmarks barely drops even at high watermark strength.
Nonetheless, our analysis of reasoning traces tells a different story.
Watermarks may severely degrade the reasoning of the models (\eg hallucinate facts, misinterpret data) and may even harm their visual understanding.
For reasoning models in particular, we suggest watermarking only the final answer, which we show prevents such degradation.

\paragraph{Limitations}
Both benchmarks are multiple-choice and single-turn, so they do not capture open-ended or interactive 
clinical workflows such as report writing, patient communication, multi-turn dialogue, or retrieval-augmented
and tool-augmented reasoning. In such settings, each watermarked response becomes context for subsequent generation, and it
remains unclear whether watermarking-induced degradation compounds or attenuates across turns. Future work should extend the evaluation to these settings.
Additionally, our evaluation is restricted to open-weight models (watermarking closed-source models is, in most cases, not possible). 
Because proprietary models are often substantially larger, they may behave differently under watermarking.
Finally, our judge pipeline, while validated against two board-certified physicians on a subset of the data, must reliably detect even subtle failure modes (\eg hallucinated image findings), which requires a judge (LLM or VLM) substantially more capable than the models being evaluated. 

\paragraph{Future work: improving watermark design.}
The watermarking methods we test are model- and domain-agnostic and therefore do not account for the
clinical importance of specific tokens. 
It remains unclear whether domain-adapted schemes would outperform generic ones or how such schemes should be designed in practice.
Given the societal impacts, we believe future work should explore watermarking methods that better preserve clinically critical content.

\message{^^JLASTBODYPAGE \thepage^^J}

\clearpage

\begin{ack}
  Robin Staab has been supported by the SERI grant SAFEAI (Certified Safe, Fair and Robust Artificial Intelligence,
contract no. MB22.00088). Views and opinions expressed are
however those of the authors only and do not necessarily reflect
those of the European Union or European Commission. Neither
the European Union nor the European Commission can be held
responsible for them. The work has received funding from the
Swiss State Secretariat for Education, Research and Innovation
(SERI) (SERI-funded ERC Consolidator Grant).
	We thank Dr. Andrea Radebold (Internal Medicine, Quinnipiac Medical of Branford/Yale School of Medicine) for her contribution to data annotation.
\end{ack}


\bibliographystyle{unsrtnat}
\bibliography{references}
\vfill
\clearpage

\message{^^JLASTREFERENCESPAGE \thepage^^J}

\ifincludeappendixx
	\newpage
	\appendix
	\onecolumn
	\crefalias{section}{appendix}
	\crefalias{subsection}{appendix}
	\section{Additional Experimental Details}
\label{app:additional-experimental-details}

\subsection{Watermarking Scheme Details}
\label{app:watermark-details}

All five schemes share a common structure: at each token position, a hash of the preceding context seeds a pseudorandom number generator that assigns a score~$g_u$ to every vocabulary entry~$u$. The \emph{sampling mechanism} then maps the original next-token distribution~$p$ and the scores~$g$ to a watermarked next-token distribution~$q(g)$, from which the next token is drawn. The schemes differ in the score distribution, in this mapping, and consequently in their detectability--quality trade-offs. We refer to~\citet{gloaguen2026unifiedframeworkllmwatermarks} for a unified optimization-theoretic treatment of several of the schemes below. In all cases the score assignment and the reweighting are restricted to the top-$k$ of~$p$ as configured in the vLLM sampling parameters.

\paragraph{KGW~\citep{kgw1}.}
The context-dependent hash partitions the top-$k$ vocabulary into a \emph{green list} and a \emph{red list} via i.i.d.\ Bernoulli($\tfrac{1}{2}$) draws ($g_u \in \{0,1\}$), so that in expectation half of the tokens are green. A constant logit bias~$\delta$ is added to the green-list tokens before sampling, yielding the soft logit-bias watermark
\begin{equation}
  q(g) \;\propto\; p\,\exp(\delta\, g).
\end{equation}
This is the soft Kirchenbauer scheme; we do not use the hard variant that excludes red-list tokens entirely. Detection tests whether the proportion of green-list tokens among the generated text exceeds the $\tfrac{1}{2}$ chance level via a one-sided binomial test (and a $z$-test for cross-checking). Watermark strength is controlled by~$\delta$.

\paragraph{DipMark~\citep{dipmark}.}
A distribution-preserving scheme parameterized by $\alpha\in(0,\tfrac{1}{2}]$. The context hash seeds a uniform-random permutation~$\pi$ of the top-$k$ vocabulary; tokens are then reweighted according to the symmetric $\alpha$-shift of the permuted CDF. Concretely, in the permuted ordering the scheme averages two indicator-style reweights at the boundaries $\alpha$ and $1-\alpha$ of the cumulative mass, which at $\alpha = \tfrac{1}{2}$ yields exact distortion-freeness in expectation, $\mathbb{E}_G[q(G)] = p$, and as $\alpha\!\to\!0$ progressively biases mass toward the upper half of the permutation. Detection compares each generated token's rank under the recomputed permutation against the threshold $\gamma$: tokens above the threshold count as ``green'', and we apply a one-sided binomial test (and a $z$-test) against $1-\gamma$. We evaluate only the distortion-free configuration $\alpha = \gamma = 0.5$, as is standard in the literature.

\paragraph{AAR~\citep{aar}.}
For each token position, i.i.d.\ Gumbel-distributed scores~$g_u$ (location~0, scale $\beta = 1$) are drawn and the next token is selected deterministically as
\begin{equation}
  \hat{u} \;=\; \arg\max_{u}\Bigl(g_u + \tau\,\log p_u\Bigr),\qquad \tau \;=\; \frac{\beta}{1+\beta\,\delta},
\end{equation}
and represented as a Dirac distribution at~$\hat{u}$. We fix $\beta = 1$ throughout and sweep~$\delta\geq 0$ as the watermark-strength knob. At $\delta = 0$ we recover the canonical AAR sampler $\arg\max_u(g_u + \log p_u)$: by the Gumbel-max trick the marginal distribution of~$\hat{u}$ over the score draw is exactly~$p$, so the scheme is distortion-free, and AAR is the optimal sampling mechanism among all distortion-free watermarks that use a sum-based detector~\citep{gloaguen2026unifiedframeworkllmwatermarks}, corresponding to the soft KL constraint $\mathrm{KL}(\mathbb{E}_G[q(G)]\,\|\,p) = 0$. For $\delta > 0$, $\tau < 1$ down-weights $\log p$ relative to the Gumbel score, biasing the argmax toward the score sequence and yielding a more easily detectable but no-longer-distortion-free signal.

\paragraph{PPL~\citep{gloaguen2026unifiedframeworkllmwatermarks}.}
For each token position, the scheme draws i.i.d.\ Bernoulli-sum scores $g_u\sim\mathrm{Binomial}(N,\tfrac{1}{2})$ and emits a Dirac distribution at the \emph{tilted argmax}
\begin{equation}
  \hat{u}(\beta) \;=\; \arg\max_u \bigl(g_u + \beta\,\log p_u\bigr),
\end{equation}
where the tilt~$\beta\geq 0$ is solved per token by 1D bisection so that the expected log-likelihood of~$\hat{u}$ exactly saturates the per-token log-perplexity budget,
\begin{equation}
  \mathbb{E}_G\bigl[\log p_{\hat{u}(\beta)}\bigr] \;=\; \sum_v p_v\log p_v \;-\; \varepsilon.
\end{equation}
The constraint is the same per-token log-perplexity budget $(p - q(g))\cdot\log p \leq \varepsilon$ as the \emph{hard} variant from the unified framework, but enforced \emph{in expectation over the scores} via the scalar tilt~$\beta$ rather than by a per-realization linear program; in practice this matches the constraint while admitting the simple Gumbel-style implementation. The expectation in the bisection is approximated by 128 Monte Carlo score draws, with a fixed schedule of 60 bisection iterations and the search bracket $\beta\in[0,100]$. Watermark strength is controlled by~$\varepsilon$ (larger~$\varepsilon$ permits more deviation from the model's distribution and thus a stronger watermark).

\paragraph{SynthID~\citep{synthid}.}
For each token, $m$~independent layers of i.i.d.\ Bernoulli scores $g^{(i)} \in \{0,1\}^{|V|}$ are drawn (in our configuration $m = 30$). Starting from $q^{(0)} := p$, the algorithm applies $m$~tournament reweighting layers parameterized by the number of leaves per match~$\ell$, controlling the strength of the watermark. With $\ell = 2$ (\emph{non-distortionary} tournament), each layer applies the multiplicative update
\begin{equation}
  q^{(i)}_u \;=\; q^{(i-1)}_u\bigl(1 + g^{(i)}_u - q^{(i-1)}\!\cdot g^{(i)}\bigr),
\end{equation}
which preserves the marginal distribution $\mathbb{E}_{g^{(i)}}[q^{(i)}] = q^{(i-1)}$. With $\ell > 2$ leaves per match (\emph{distortionary} tournament), the layer instead applies the closed-form tournament-winner reweight
\begin{equation}
  q^{(i)}_u \;=\; q^{(i-1)}_u \cdot
  \begin{cases}
    \dfrac{1 - (1 - m_i)^{\ell}}{m_i} & \text{if } g^{(i)}_u = 1,\\[6pt]
    (1 - m_i)^{\ell - 1} & \text{if } g^{(i)}_u = 0,
  \end{cases}\qquad m_i := q^{(i-1)}\!\cdot g^{(i)},
\end{equation}
which biases each layer toward tokens with $g^{(i)}_u = 1$ and is no longer distortion-free for $\ell > 2$. Stacking layers progressively strengthens the watermark signal. The original SynthID-Text scheme of~\citet{synthid} is typically configured with $\ell = 2$ leaves (the non-distortionary tournament) and a depth in the range $m \in \{20, 30\}$, with detectability controlled by varying~$m$.

\subsection{Generation and Reproducibility}
\label{app:generation_reproducibility}

All completions are generated with vLLM (v0.19.1), with the watermark realised as a custom \texttt{LogitsProcessor} so that watermarked and unwatermarked runs differ only at the logit-modification step. All experiments use the model-native tokenizer without post-processing. Decoding uses temperature $0.7$ and unrestricted top-$k$ throughout. Each (model, watermark, strength) configuration is run with a fixed sampling \texttt{seed} (default $0$); paired watermarked/unwatermarked completions for a given question therefore share both prompt and seed, isolating the watermark as the only source of divergence. The watermark uses output-only context hashing, so the first $4$ generated tokens (one context window) are not watermarked; this affects $<\!1\%$ of tokens and is consistent between generation and detection.

For non-reasoning models we cap output at $2{,}500$ tokens, while for reasoning models we allocate a generation budget of $10{,}000$ tokens, and post-hoc split out the answer portion (after \texttt{</think>}) with a maximum of $2{,}500$ tokens for comparability with the non-reasoning pipeline; the watermark is applied to both reasoning trace and answer tokens by default. In practice, the $2{,}500$-token cap never truncated a well-formed response; the only completions reaching the limit were degenerate outputs (e.g.\ repetitive token loops), which are captured by the format/termination failure axis of the single-response audit. Answer letters are extracted by a two-tier regex parser. A primary pass scans for structured formats observed across models — \texttt{ANSWER: X}, \texttt{FINAL ANSWER: X}, \texttt{ANSWER: (X)}, \texttt{ANSWER: <X>}, \texttt{\textbackslash boxed\{X\}}, \texttt{FINAL ANSWER is X}, and \texttt{FINAL ANSWER: \textit{<free text>} (X)} where exactly one letter appears parenthesised. If, and only if, the primary pass finds nothing, a fallback pass admits conservative natural-language patterns (e.g.\ ``the answer is X'', ``I would choose X'', or a candidate name followed by a parenthesised letter at end of line). The parser only commits to a letter when exactly one option ($\mathrm{A}$--$\mathrm{E}$) is unambiguously indicated; otherwise the response is recorded as unanswered.

All experiments run on a single node with up to $8$ NVIDIA RTX Blackwell $6000$ Pro GPUs ($96$\,GB VRAM each). Watermark implementations, generation and analysis pipelines are released alongside the paper to enable reproduction of the reported numbers.

\subsection{Existing Asset Licenses}
\label{app:asset_licenses}

We use existing datasets and model checkpoints only for evaluation, and cite the original creators in \cref{sec:experimental_setup}. The licenses below are the license names stated by the corresponding official repositories or model cards.

\paragraph{Datasets.}
\begin{itemize}[leftmargin=1.4em]
    \item \emph{MedQA}: MIT License.
    \item \emph{MedXpertQA-MM}: MIT License.
\end{itemize}

\paragraph{Models.}
\begin{itemize}[leftmargin=1.4em]
    \item \emph{\textsc{Llama-3.1-8B}} and \emph{\textsc{Llama-3.1-70B}}: Meta Llama 3 Community License Agreement.
    \item \emph{\textsc{BioMistral-7B}}: Apache License 2.0.
    \item \emph{\textsc{Gemma-3-12B}}: Gemma Terms of Use.
    \item \emph{\textsc{Phi-4-14B}}: MIT License.
    \item \emph{\textsc{OpenBioLLM-70B}} and \emph{\textsc{UltraMedical-70B}}: Meta Llama 3 Community License Agreement.
    \item \emph{\textsc{DeepSeek-R1-Distill-Llama-8B}}, \emph{\textsc{DeepSeek-R1-Distill-Llama-70B}}, and \emph{\textsc{DeepSeek-R1-Distill-Qwen-32B}}: MIT License; the Llama-derived variants also inherit the applicable Meta Llama Community License terms, and the Qwen-derived variant inherits the applicable Apache License 2.0 upstream terms.
    \item \emph{\textsc{Qwen-3.5-27B}}, \emph{\textsc{Qwen-3-VL-8B}}, and \emph{\textsc{Qwen-3-VL-32B}}: Apache License 2.0.
    \item \emph{\textsc{Gemma-4-31B}}: Apache License 2.0.
    \item \emph{\textsc{Lingshu-7B}} and \emph{\textsc{Lingshu-32B}}: MIT License.
    \item \emph{\textsc{MedGemma-4B}} and \emph{\textsc{MedGemma-27B}}: Health AI Developer Foundations Terms of Use.
\end{itemize}

\section{Per-Configuration Accuracy and Perplexity}
\label{app:acc_tables}
\cref{tab:medqa_acc_all,tab:medxpertqa_mm_acc_all} report detectability (TPR at $1\%$ FPR) and benchmark accuracy for every (model, scheme, strength) triple evaluated in \cref{sec:accuracy}, one benchmark per table. Rows are grouped by scheme and sorted by increasing watermark strength; each model contributes a (TPR, Acc) sub-column pair, with the unwatermarked baseline shown beneath the model header. Configurations strictly stronger than the first to reach TPR\,$\geq 99\%$ are omitted (---), as they yield no additional detectability.


\begin{sidewaystable}[p]
\centering
\caption{MedQA: watermark detectability (TPR @ 1\%FPR, in \%) and benchmark accuracy (Acc, in \%) for every (model, scheme, strength) configuration evaluated in \cref{sec:accuracy}, computed over $N=2000$ questions (full benchmark). Per-model unwatermarked baseline accuracy is shown beneath each model header. For each (model, scheme) we report all configurations up to and including the first one with TPR\,$\geq$\,99\%; stronger configurations beyond that saturation point are omitted (---), as they yield no additional detectability.}
\label{tab:medqa_acc_all}
\footnotesize
\setlength{\tabcolsep}{4pt}
\renewcommand{\arraystretch}{1.05}
\begin{tabular}{ll *{6}{rr}}
\toprule
 & & \multicolumn{2}{c}{\makecell{\textsc{Gemma-3-12B}\\(68.2)}}
   & \multicolumn{2}{c}{\makecell{\textsc{Llama-3.1-70B}\\(91.0)}}
   & \multicolumn{2}{c}{\makecell{\textsc{Llama-3.1-8B}\\(72.8)}}
   & \multicolumn{2}{c}{\makecell{\textsc{OpenBioLLM-70B}\\(73.8)}}
   & \multicolumn{2}{c}{\makecell{\textsc{Phi-4-14B}\\(81.5)}}
   & \multicolumn{2}{c}{\makecell{\textsc{UltraMedical-70B}\\(86.6)}} \\
\cmidrule(lr){3-4} \cmidrule(lr){5-6} \cmidrule(lr){7-8} \cmidrule(lr){9-10} \cmidrule(lr){11-12} \cmidrule(lr){13-14}
Scheme & Strength
 & TPR & Acc & TPR & Acc & TPR & Acc & TPR & Acc
 & TPR & Acc & TPR & Acc \\
\midrule
AAR & $\delta=0$    & 96.3 & 69.2 & 37.3 & 89.5 & 91.4 & 68.5 & 36.6 & 76.0 & 70.0 & 82.5 & 67.8 & 87.0 \\
AAR & $\delta=0.1$  & 98.2 & 68.5 & 47.1 & 90.1 & 95.7 & 68.9 & 45.3 & 74.5 & ---  & ---  & 76.5 & 86.9 \\
AAR & $\delta=0.2$  & 99.3 & 69.2 & 55.6 & 89.6 & 98.1 & 65.8 & 51.7 & 73.7 & 86.0 & 81.3 & 84.4 & 86.8 \\
AAR & $\delta=0.3$  & ---  & ---  & 64.7 & 90.8 & 99.4 & 62.0 & 59.1 & 74.9 & 92.0 & 81.6 & 90.5 & 86.2 \\
AAR & $\delta=0.4$  & ---  & ---  & 73.1 & 89.5 & ---  & ---  & 66.7 & 72.7 & 96.2 & 80.8 & 93.5 & 86.5 \\
AAR & $\delta=0.5$  & ---  & ---  & 80.7 & 89.8 & ---  & ---  & 76.1 & 72.0 & 98.1 & 79.7 & 95.6 & 85.4 \\
\midrule
DipMark & $\alpha=0.5$ & 90.1 & 68.5 & 30.2 & 89.8 & 83.8 & 67.2 & 25.4 & 72.0 & 59.8 & 81.1 & 57.7 & 87.5 \\
\midrule
KGW & $\delta=0.5$ & 16.9 & 69.2 &  6.4 & 90.1 & 18.6 & 70.5 &  3.1 & 74.6 &  5.9 & 81.8 & 11.7 & 87.6 \\
KGW & $\delta=1$   & 65.1 & 69.7 & 22.2 & 90.0 & 63.0 & 69.6 & 11.4 & 74.9 & 27.5 & 82.2 & 41.2 & 86.7 \\
KGW & $\delta=2$   & 99.4 & 68.8 & 70.7 & 89.6 & 99.2 & 64.5 & 51.4 & 73.4 & 87.3 & 80.7 & 91.4 & 85.9 \\
KGW & $\delta=3$   & ---  & ---  & 95.2 & 90.0 & ---  & ---  & 85.5 & 69.0 & 99.2 & 80.3 & 99.1 & 85.2 \\
KGW & $\delta=4$   & ---  & ---  & 99.1 & 90.4 & ---  & ---  & 95.9 & 63.2 & ---  & ---  & ---  & ---  \\
KGW & $\delta=5$   & ---  & ---  & ---  & ---  & ---  & ---  & 98.8 & 54.9 & ---  & ---  & ---  & ---  \\
\midrule
PPL & $\varepsilon=0$   & 99.5 & 68.4 & 81.7 & 89.6 & 99.1 & 68.3 & 57.2 & 73.4 & 90.8 & 78.8 & 94.4 & 80.2 \\
PPL & $\varepsilon=0.1$ & ---  & ---  & 92.8 & 88.2 & ---  & ---  & 73.9 & 73.5 & 96.7 & 78.8 & 97.7 & 78.5 \\
PPL & $\varepsilon=0.2$ & ---  & ---  & 97.8 & 89.3 & ---  & ---  & 86.6 & 71.9 & ---  & ---  & 99.3 & 70.0 \\
PPL & $\varepsilon=0.3$ & ---  & ---  & 99.3 & 87.9 & ---  & ---  & 92.7 & 69.8 & 99.8 & 77.5 & ---  & ---  \\
PPL & $\varepsilon=0.4$ & ---  & ---  & ---  & ---  & ---  & ---  & 95.4 & 69.2 & ---  & ---  & ---  & ---  \\
PPL & $\varepsilon=0.5$ & ---  & ---  & ---  & ---  & ---  & ---  & 97.5 & 67.5 & ---  & ---  & ---  & ---  \\
\midrule
SynthID & $\ell=2$ & 91.1 & 69.5 & 38.3 & 89.8 & 85.9 & 71.7 & 22.9 & 75.9 & 48.4 & 81.2 & 67.4 & 87.0 \\
SynthID & $\ell=3$ & 98.9 & 69.5 & 63.9 & 89.6 & 98.3 & 65.0 & 48.1 & 73.9 & 84.5 & 80.8 & 89.2 & 84.0 \\
SynthID & $\ell=4$ &100.0 & 68.8 & 79.4 & 89.6 &100.0 & 48.4 & 70.5 & 71.2 & 97.2 & 80.0 & 96.3 & 81.2 \\
SynthID & $\ell=5$ & ---  & ---  & 87.9 & 90.4 & ---  & ---  & 83.5 & 67.2 & 99.1 & 79.3 & 98.8 & 76.8 \\
SynthID & $\ell=6$ & ---  & ---  & 93.4 & 90.3 & ---  & ---  & 93.3 & 59.2 & ---  & ---  & 99.2 & 63.1 \\
\bottomrule
\end{tabular}
\end{sidewaystable}


\begin{sidewaystable}[p]
\centering
\caption{MedXpertQA-MM: watermark detectability (TPR @ 1\%FPR, in \%) and benchmark accuracy (Acc, in \%) for every (model, scheme, strength) configuration evaluated in \cref{sec:accuracy}, computed over $N=2000$ questions (full benchmark). Per-model unwatermarked baseline accuracy is shown beneath each model header. For each (model, scheme) we report all configurations up to and including the first one with TPR\,$\geq$\,99\%; stronger configurations beyond that saturation point are omitted (---), as they yield no additional detectability.}
\label{tab:medxpertqa_mm_acc_all}
\footnotesize
\setlength{\tabcolsep}{4pt}
\renewcommand{\arraystretch}{1.05}
\begin{tabular}{ll *{7}{rr}}
\toprule
 & & \multicolumn{2}{c}{\makecell{\textsc{Gemma-4-31B}\\(54.9)}}
   & \multicolumn{2}{c}{\makecell{\textsc{Lingshu-32B}\\(29.9)}}
   & \multicolumn{2}{c}{\makecell{\textsc{Lingshu-7B}\\(24.3)}}
   & \multicolumn{2}{c}{\makecell{\textsc{MedGemma-27B}\\(32.8)}}
   & \multicolumn{2}{c}{\makecell{\textsc{MedGemma-4B}\\(24.4)}}
   & \multicolumn{2}{c}{\makecell{\textsc{Qwen-3-VL-32B}\\(38.9)}}
   & \multicolumn{2}{c}{\makecell{\textsc{Qwen-3-VL-8B}\\(29.8)}} \\
\cmidrule(lr){3-4} \cmidrule(lr){5-6} \cmidrule(lr){7-8} \cmidrule(lr){9-10} \cmidrule(lr){11-12} \cmidrule(lr){13-14} \cmidrule(lr){15-16}
Scheme & Strength
 & TPR & Acc & TPR & Acc & TPR & Acc & TPR & Acc
 & TPR & Acc & TPR & Acc & TPR & Acc \\
\midrule
AAR & $\delta=0$    & 65.2 & 55.4 & 72.6 & 31.4 & 73.7 & 23.8 & 90.0 & 33.6 & 84.2 & 24.2 & 93.5 & 39.1 & 97.9 & 26.5 \\
AAR & $\delta=0.1$  & 71.2 & 55.0 & 82.5 & 29.6 & 82.9 & 23.5 & 94.2 & 34.2 & 91.2 & 24.7 & 95.5 & 39.3 & 98.6 & 27.0 \\
AAR & $\delta=0.2$  & 79.5 & 54.6 & 89.5 & 28.8 & 91.4 & 23.4 & 96.6 & 33.0 & 94.0 & 24.8 & 97.9 & 40.4 & 99.1 & 27.3 \\
AAR & $\delta=0.3$  & 84.8 & 54.7 & 93.8 & 30.1 & 94.5 & 24.2 & 98.0 & 33.5 & 96.8 & 24.0 & 98.9 & 39.0 & ---  & ---  \\
AAR & $\delta=0.4$  & 90.5 & 54.1 & 96.2 & 30.4 & 97.5 & 23.8 & 98.9 & 32.1 & 98.2 & 23.9 & 99.5 & 38.7 & ---  & ---  \\
AAR & $\delta=0.5$  & 92.8 & 53.6 & 98.3 & 30.2 & 98.5 & 23.5 & 99.6 & 31.9 & 99.0 & 22.2 & ---  & ---  & ---  & ---  \\
\midrule
DipMark & $\alpha=0.5$ & 45.6 & 54.4 & 57.0 & 31.2 & 57.5 & 25.7 & 82.0 & 32.4 & 72.5 & 23.6 & 87.1 & 39.0 & 94.5 & 29.0 \\
\midrule
KGW & $\delta=0.5$ &  7.1 & 54.3 &  9.2 & 30.6 & 10.1 & 24.9 & 14.8 & 34.1 & 14.1 & 22.1 & 17.1 & 38.8 & 30.3 & 27.8 \\
KGW & $\delta=1$   & 26.9 & 54.0 & 34.1 & 31.8 & 38.9 & 23.4 & 59.2 & 32.5 & 54.0 & 23.2 & 62.3 & 40.1 & 81.2 & 29.1 \\
KGW & $\delta=2$   & 81.0 & 54.4 & 91.0 & 29.5 & 91.0 & 22.4 & 97.0 & 33.6 & 91.8 & 23.8 & 98.0 & 39.7 & 99.2 & 29.2 \\
KGW & $\delta=3$   & 96.9 & 54.4 & 99.2 & 30.4 & 99.0 & 25.2 &100.0 & 32.5 & 96.5 & 21.5 & 99.9 & 38.5 & ---  & ---  \\
KGW & $\delta=4$   & 99.9 & 54.0 & ---  & ---  & ---  & ---  & ---  & ---  & ---  & ---  & ---  & ---  & ---  & ---  \\
KGW & $\delta=5$   & ---  & ---  & ---  & ---  & ---  & ---  & ---  & ---  & 97.2 & 21.8 & ---  & ---  & ---  & ---  \\
\midrule
PPL & $\varepsilon=0$   & 88.9 & 53.3 & 92.8 & 30.6 & 94.7 & 23.9 & 98.6 & 33.1 & 91.1 & 22.4 & 99.1 & 37.4 & 99.6 & 26.9 \\
PPL & $\varepsilon=0.1$ & 96.1 & 52.6 & 97.8 & 29.6 & 97.4 & 23.2 & 99.7 & 32.5 & ---  & ---  & ---  & ---  & ---  & ---  \\
PPL & $\varepsilon=0.2$ & 98.8 & 53.8 & 99.4 & 29.2 & 99.2 & 23.9 & ---  & ---  & 94.0 & 21.9 & ---  & ---  & ---  & ---  \\
PPL & $\varepsilon=0.3$ & 99.6 & 53.4 & ---  & ---  & ---  & ---  & ---  & ---  & ---  & ---  & ---  & ---  & ---  & ---  \\
PPL & $\varepsilon=0.4$ & ---  & ---  & ---  & ---  & ---  & ---  & ---  & ---  & 96.4 & 21.1 & ---  & ---  & ---  & ---  \\
PPL & $\varepsilon=0.5$ & ---  & ---  & ---  & ---  & ---  & ---  & ---  & ---  & 96.7 & 21.1 & ---  & ---  & ---  & ---  \\
\midrule
SynthID & $\ell=2$ & 61.2 & 53.8 & 63.3 & 30.4 & 69.0 & 23.8 & 92.9 & 31.9 & 88.6 & 24.5 & 89.0 & 40.0 & 94.8 & 26.7 \\
SynthID & $\ell=3$ & 81.2 & 53.8 & 89.6 & 30.3 & 92.5 & 24.1 & 98.7 & 32.9 & 95.0 & 22.6 & 98.1 & 38.6 & 99.4 & 26.4 \\
SynthID & $\ell=4$ & 90.7 & 53.8 & 98.0 & 27.9 & 98.0 & 24.3 & 99.9 & 32.6 & 96.5 & 21.6 & 99.5 & 39.0 & ---  & ---  \\
SynthID & $\ell=5$ & 95.4 & 52.9 & 99.4 & 26.6 & 99.8 & 23.0 & ---  & ---  & 96.6 & 20.1 & ---  & ---  & ---  & ---  \\
\bottomrule
\end{tabular}
\end{sidewaystable}

\begin{table}[htbp]
  \centering
  \footnotesize
  \setlength{\tabcolsep}{4pt}
  \begin{tabular}{@{}l l r r r@{}}
    \toprule
    Model & Condition & Acc.\ (\%) & TPR (\%) & Mean output length (chars) \\
    \midrule
    \rowcolor{black!8} \multirow{7}{*}{\textsc{DeepSeek-R1-Distill-Llama-8B}}
      & No WM                  & 59.2 &   --  &  9{,}023 \\
      & FA-only, $\delta{=}1$  & 61.0 &  14.3 &  8{,}890 \\
      & FA-only, $\delta{=}3$  & 59.2 &  87.1 &  8{,}856 \\
      & FA-only, $\delta{=}5$  & 58.9 &  96.8 &  8{,}870 \\
      & Full WM, $\delta{=}1$  & 61.2 &  15.4 &  9{,}165 \\
      & Full WM, $\delta{=}3$  & 59.1 &  93.0 & 11{,}621 \\
      & Full WM, $\delta{=}5$  & 57.2 &  98.9 & 15{,}257 \\
    \cmidrule(lr){1-5}
    \rowcolor{black!8} \multirow{7}{*}{\textsc{DeepSeek-R1-Distill-Qwen-32B}}
      & No WM                  & 83.6 &   --  &  6{,}447 \\
      & FA-only, $\delta{=}1$  & 83.6 &   8.1 &  6{,}362 \\
      & FA-only, $\delta{=}3$  & 82.9 &  74.3 &  6{,}444 \\
      & FA-only, $\delta{=}5$  & 81.5 &  97.7 &  6{,}482 \\
      & Full WM, $\delta{=}1$  & 84.1 &  10.3 &  6{,}737 \\
      & Full WM, $\delta{=}3$  & 83.7 &  84.4 &  8{,}166 \\
      & Full WM, $\delta{=}5$  & 82.1 &  99.6 & 10{,}027 \\
    \cmidrule(lr){1-5}
    \rowcolor{black!8} \multirow{7}{*}{\textsc{DeepSeek-R1-Distill-Llama-70B}}
      & No WM                  & 92.3 &   --  &  6{,}025 \\
      & FA-only, $\delta{=}1$  & 92.4 &   7.0 &  5{,}995 \\
      & FA-only, $\delta{=}3$  & 91.9 &  59.2 &  6{,}072 \\
      & FA-only, $\delta{=}5$  & 90.8 &  92.4 &  6{,}088 \\
      & Full WM, $\delta{=}1$  & 91.9 &   7.1 &  6{,}239 \\
      & Full WM, $\delta{=}3$  & 91.4 &  82.3 &  7{,}227 \\
      & Full WM, $\delta{=}5$  & 89.8 &  99.7 &  8{,}759 \\
    \cmidrule(lr){1-5}
    \rowcolor{black!8} \multirow{7}{*}{\textsc{Qwen-3.5-27B}}
      & No WM                  & 94.6 &   --  & 13{,}258 \\
      & FA-only, $\delta{=}1$  & 94.1 &  23.5 & 13{,}125 \\
      & FA-only, $\delta{=}3$  & 94.5 &  97.5 & 13{,}285 \\
      & FA-only, $\delta{=}5$  & 94.1 &  98.8 & 13{,}743 \\
      & Full WM, $\delta{=}1$  & 94.8 &  26.1 & 13{,}285 \\
      & Full WM, $\delta{=}3$  & 94.0 &  99.0 & 14{,}976 \\
      & Full WM, $\delta{=}5$  & 77.1 &  99.5 & 18{,}019 \\
    \bottomrule
  \end{tabular}
  \caption{MedQA accuracy, TPR at $1\%$ FPR, and mean output length (reasoning trace plus final answer, characters) for four reasoning models under No~WM, FA-only (post-\texttt{<thinking>} tokens only), and Full~WM (entire output), KGW $\delta \in \{1, 3, 5\}$. TPR is computed on the final-answer tokens only.}
  \label{tab:thinking_acc_tpr_length}
\end{table}

\subsection{Perplexity Under Watermarking}
\label{app:ppl}

\cref{fig:ppl_medqa,fig:ppl_medxpert} plot generation perplexity against watermark detectability. Perplexity increases monotonically with watermark strength on every model and scheme, but its magnitude is a poor predictor of downstream harm: \textsc{Gemma-3-12B} shows clear perplexity shifts under KGW with no accuracy effect, while \textsc{UltraMedical-70B} suffers a $31$\,pp accuracy collapse under PPL despite only moderate perplexity shifts. Exact per-configuration values accompany the code release.

\begin{figure}[H]
    \centering
    \includegraphics[width=\textwidth]{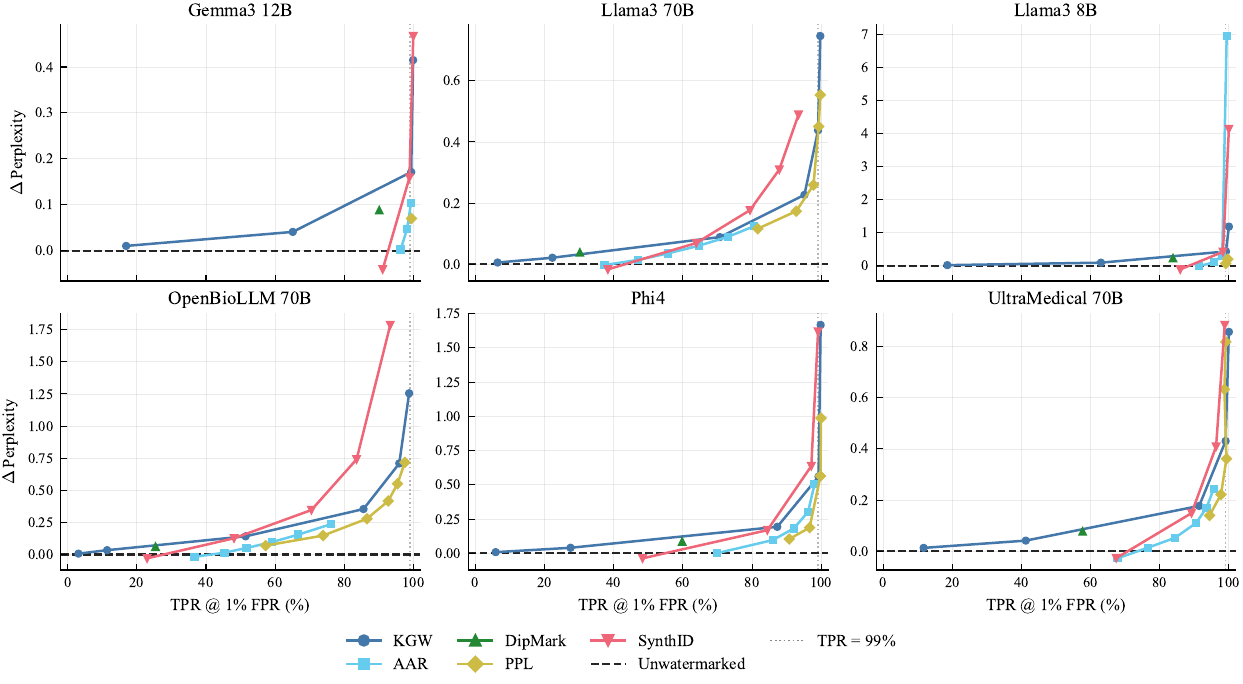}
    \caption{\textbf{MedQA: generation-perplexity shift vs.\ watermark detectability}, six non-reasoning LLMs. $\Delta$\,PPL\,$=$\,PPL$_{\mathrm{wm}}-$PPL$_{\mathrm{base}}$. Monotonic in strength on every model$\times$scheme cell; absolute magnitude spans an order of magnitude (note differing $y$-axes). \textsc{Llama-3.1-8B} has the steepest increases; \textsc{Gemma-3-12B} and \textsc{UltraMedical-70B} show contained shifts despite their sharply different accuracy robustness.}
    \label{fig:ppl_medqa}
\end{figure}

\begin{figure}[H]
    \centering
    \includegraphics[width=\textwidth]{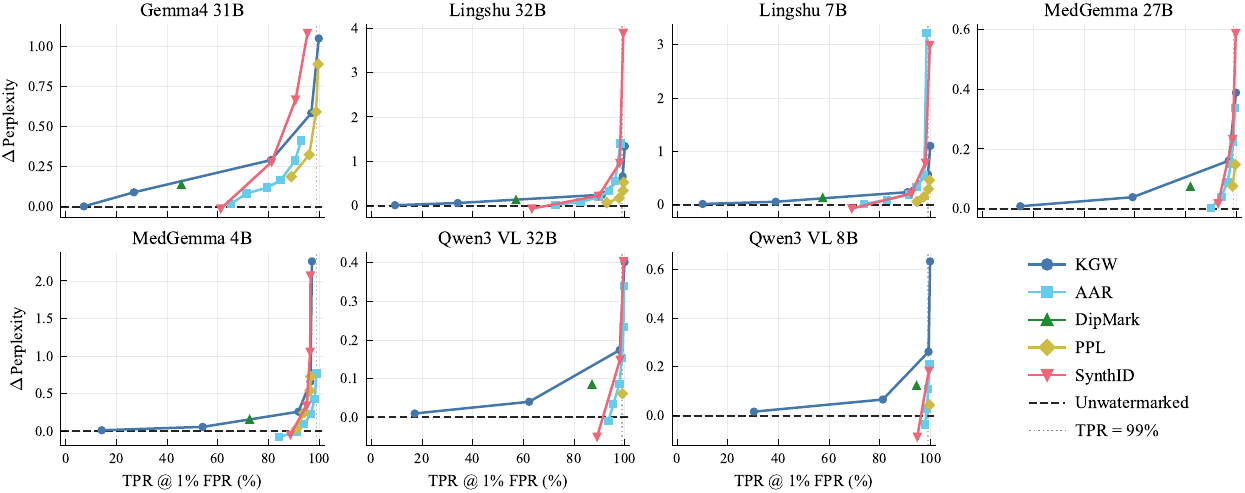}
    \caption{\textbf{MedXpertQA-MM: generation-perplexity shift vs.\ watermark detectability}, seven VLMs. Same overall shape as MedQA; KGW and SynthID drive the steepest high-TPR increases. \textsc{Lingshu-32B} and \textsc{Lingshu-7B} show pronounced jumps at the highest strengths despite minimal accuracy changes (cf.\ \cref{fig:tpr_acc_medxpert}).}
    \label{fig:ppl_medxpert}
\end{figure}

\section{Audit Outcomes Across Watermark Detectability}
\label{app:dose_response}

\cref{sec:reasoning_degradation} fixes a single deployment operating point per cell ($\geq 99\%$ TPR), leaving the \emph{shape} of the strength--harm relationship unresolved -- whether harm grows smoothly with detectability, or sits near baseline up to a threshold and then jumps.
We sweep watermark strength on two contrasting cells: \emph{\textsc{Phi-4-14B}} on MedQA (an accuracy-stable text-only LLM) and \emph{\textsc{MedGemma-4B}} on MedXpertQA-MM (a smaller medical VLM), under the two schemes whose $\geq 99\%$-TPR signatures differ most in \cref{tab:main_combined} -- \emph{KGW} (broad multi-axis effect) and \emph{SynthID} (the largest silent-error cell on both models).

\cref{fig:dose_response} plots one harm metric per axis-of-interest against detectability. For \textsc{Phi-4-14B} we report the per-response confident-error rate (a major component of F\&C) and the pairwise vignette-fabrication rate (the silent-error signature picked up by VigFab). For \textsc{MedGemma-4B} we report the net visual signal used in \cref{fig:image_decomp} and again vignette-fabrication, so the silent-error axis is comparable across the two cells.

\paragraph{KGW vs.\ SynthID: monotonic slider vs.\ threshold cliff.}
KGW degrades smoothly with detectability. On \textsc{Phi-4-14B}, watermarked confident-error climbs from $14.2\%$ at TPR $5.9\%$ to $21.6\%$ at TPR $99.9\%$, against a stable $\approx 12\%$ unwatermarked baseline; on \textsc{MedGemma-4B}, the net visual signal slides from $+0.09$ at TPR $14\%$ through zero to $-0.21$ at TPR $97\%$ (baseline $\approx +0.10$).
SynthID is threshold-y: it tracks baseline up to TPR $\approx 90\%$ and then jumps in the final detectability decade. On \textsc{Phi-4-14B}, confident-error is flat at $\approx 13.7\%$ across TPRs $48$--$84\%$, then rises to $19.1\%$ (TPR $97\%$) and $25.0\%$ (TPR $99\%$); pairwise vignette-fabrication stays $\leq 0.8\%$ up to TPR $97\%$ and jumps to $3.8\%$ at TPR $99\%$. On \textsc{MedGemma-4B}, vignette-fabrication tracks $5.8\% \to 7.9\% \to 18.7\%$ at TPRs $89, 95, 97\%$.

These shapes have different operational consequences: a KGW deployment can move a slider to trade detectability against quality, whereas a SynthID deployment at the detectability a regulator would require sits past the elbow of a cliff.

\begin{figure}[t]
    \centering
    \includegraphics[width=\textwidth]{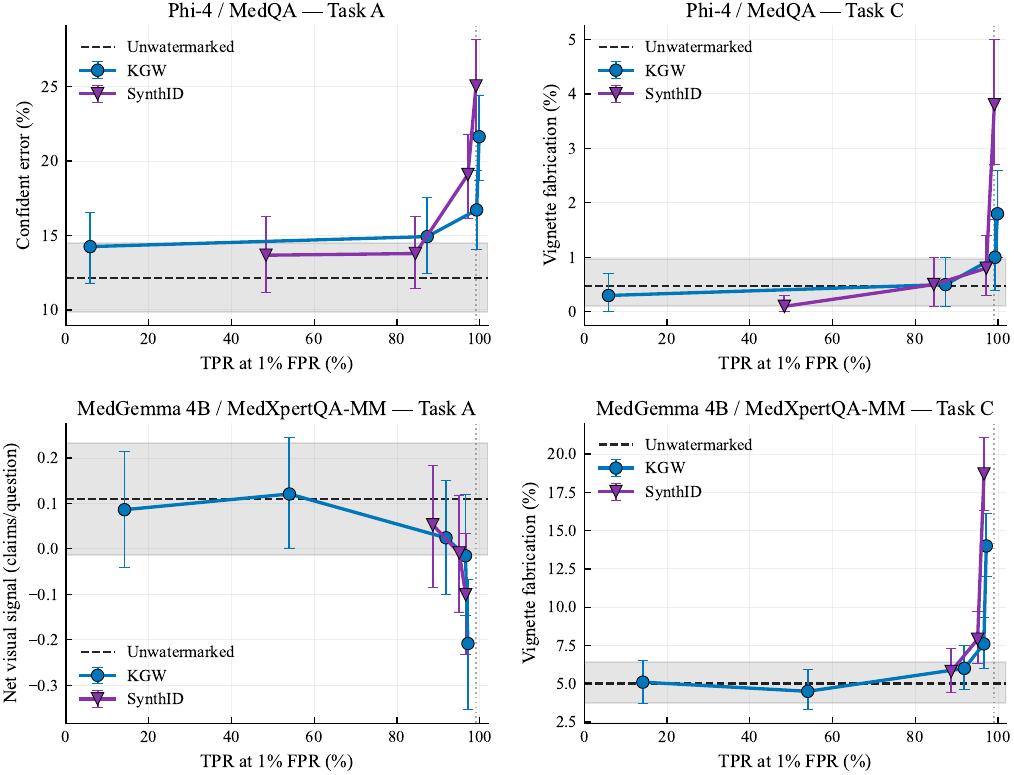}
    \caption{\textbf{Audit outcomes vs.\ watermark detectability} on \textsc{Phi-4-14B} (MedQA, top) and \textsc{MedGemma-4B} (MedXpertQA-MM, bottom) under KGW (blue) and SynthID (purple). Each point is one (scheme, strength); dashed line and grey band give the unwatermarked baseline mean and $95\%$-bootstrap CI; dotted vertical marks TPR\,$=99\%$. Top: per-response confident-error rate restricted to right-letter responses (left); pairwise vignette-fabrication rate (right). Bottom: net visual signal, $\textsc{supported}-\textsc{contradicted}$ visual claims per question (left); pairwise vignette-fabrication rate (right).}
    \label{fig:dose_response}
\end{figure}

\subsection{Full Reasoning-Audit Grids}
\label{app:full_judge_tables}

\cref{sec:reasoning_degradation} reports only the two most-affected models per benchmark. Below: the full $7$-model grids per benchmark, per-scheme FabRatio breakdowns, and the all-VLM companion to \cref{fig:image_decomp}.

\begin{sidewaystable}
\centering
\scriptsize
\setlength{\tabcolsep}{1.5pt}
\begin{threeparttable}
\caption{\textbf{Full reasoning audit at $99\%$ TPR on MedQA}, for all $7$ LLMs.
We report the accuracy ($\Delta$Acc), the confident-error rate ($\Delta$ConfErr), the faulty-but-correct rate ($\Delta$F\&C), the percentage of corrupted spellings ($\Delta$Spell/100), the percentage of fabricated facts ($\Delta$Fab/100), the within-response contradiction rate ($\Delta$Contr), vignette fabrication ($\Delta$VigFab), clue misreads ($\Delta$Misread), formatting failures ($\Delta$Fmt), linguistic corruption ($\Delta$Ling), the percentage of misapplied medical terms ($\Delta$Misapp/100), the percentage of paired responses with major diagnostic divergence (Major\%), and the vignette fabrication ratio (FabRatio).
For unwatermarked rows we report the actual value, whereas for watermarked rows we report the difference/ratio with the unwatermarked baseline.
We \textbf{bold} entries that are statistically significant (p-value below $0.01$).
}\label{tab:taskA_full_MedQA}
\begin{tabular}{llrrrrrrrrrrrrr}
\toprule
Model & Scheme & $\Delta$Acc & $\Delta$ConfErr & $\Delta$F\&C & $\Delta$Spell/100 & $\Delta$Fab/100 & $\Delta$Contr & $\Delta$VigFab & $\Delta$Misread & $\Delta$Fmt & $\Delta$Ling & $\Delta$Misapp/100 & Major\% & FabRatio \\
\midrule
\rowcolor{black!8} \textsc{Llama-3.1-8B} & Unwatermarked & 73.3 & 35.3 & 29.6 & 0.8 & 1.4 & 22.6 & 5.8 & 6.6 & 0.9 & 0.1 & 14.7 & -- & 1.0$\times$ \\
 & AAR & -0.6 & \textbf{+8.1} & \textbf{+7.4} & \textbf{+5.8} & \textbf{+4.1} & \textbf{+6.1} & \textbf{+5.5} & \textbf{+3.2} & \textbf{+1.5} & \textbf{+6.1} & +0.2 & 31.9 & \textbf{2.1$\times$} \\
 & DipMark & -1.6 & -0.7 & +1.0 & +0.9 & +1.0 & +3.1 & +1.3 & \textbf{+3.4} & -0.3 & +0.5 & -1.5 & 29.2 & 1.6$\times$ \\
 & KGW & -2.3 & +3.0 & +2.4 & \textbf{+1.8} & \textbf{+2.2} & \textbf{+5.1} & +1.7 & +1.6 & +0.5 & +0.3 & -0.3 & 29.0 & 1.0$\times$ \\
 & PPL & -0.7 & +5.4 & +4.3 & \textbf{+24.6} & +1.0 & +4.2 & +0.5 & \textbf{+3.8} & +0.3 & \textbf{+3.4} & +0.7 & 29.8 & 1.6$\times$ \\
 & SynthID & -2.0 & +3.0 & +2.2 & \textbf{+2.1} & +1.5 & +4.0 & \textbf{+3.7} & +2.6 & +1.5 & +0.7 & +0.2 & 33.5 & 1.2$\times$ \\
\rowcolor{black!8} \textsc{Llama-3.1-70B} & Unwatermarked & 91.8 & 6.9 & 8.1 & 0.4 & 0.1 & 5.8 & 0.4 & 1.0 & 0.0 & 0.0 & 1.5 & -- & 1.0$\times$ \\
 & AAR & -1.2 & -0.5 & +0.1 & +0.3 & 0.0 & +0.1 & +0.1 & +0.5 & 0.0 & +0.1 & -0.3 & 4.1 & -- \\
 & DipMark & -1.2 & +1.4 & +1.5 & -0.1 & -0.1 & +1.4 & -0.2 & +0.2 & +0.2 & 0.0 & 0.0 & 4.0 & -- \\
 & KGW & -1.3 & +1.8 & +1.9 & +1.0 & 0.0 & +1.6 & +0.4 & +0.3 & +0.1 & +0.1 & +0.6 & 5.1 & 1.0$\times$ \\
 & PPL & -1.8 & \textbf{+3.3} & \textbf{+7.5} & \textbf{+89.4} & \textbf{+1.9} & +2.1 & +0.3 & \textbf{+3.5} & +0.6 & \textbf{+30.9} & +0.3 & 5.0 & -- \\
 & SynthID & -0.9 & +2.0 & +1.2 & \textbf{+6.5} & +0.2 & +0.9 & +0.4 & -0.1 & +0.1 & \textbf{+3.4} & 0.0 & 5.9 & 0.0$\times$ \\
\rowcolor{black!8} \textsc{Gemma-3-12B} & Unwatermarked & 67.0 & 29.2 & 22.5 & 0.7 & 1.3 & 20.4 & 3.4 & 4.5 & 0.0 & 0.0 & 9.0 & -- & 1.0$\times$ \\
 & AAR & +1.3 & +2.5 & +3.4 & +0.2 & +0.4 & +3.2 & -0.7 & +0.3 & +0.2 & 0.0 & +3.6 & 21.0 & 0.9$\times$ \\
 & DipMark & +1.8 & -1.0 & +1.1 & -0.3 & +0.2 & +1.7 & -0.4 & -0.7 & +0.4 & +0.1 & +1.5 & 20.9 & 0.4$\times$ \\
 & KGW & +1.5 & +0.7 & +1.7 & +0.1 & 0.0 & +1.9 & +1.0 & +0.1 & +0.1 & 0.0 & +0.8 & 22.4 & 1.3$\times$ \\
 & PPL & +1.5 & +2.5 & +3.2 & \textbf{+20.8} & +1.2 & +2.1 & +0.1 & \textbf{+3.9} & +0.6 & \textbf{+3.7} & +3.4 & 23.1 & 1.1$\times$ \\
 & SynthID & +1.9 & +3.0 & \textbf{+4.5} & \textbf{+2.6} & +0.3 & +3.0 & +1.1 & +0.5 & +0.2 & +0.1 & +1.5 & 22.9 & 1.1$\times$ \\
\rowcolor{black!8} \textsc{OpenBioLLM-70B} & Unwatermarked & 73.6 & 13.9 & 12.9 & 0.6 & 0.0 & 8.6 & 2.2 & 2.2 & 14.8 & 0.1 & 4.5 & -- & 1.0$\times$ \\
 & AAR & -0.2 & \textbf{+6.6} & \textbf{+4.4} & +0.9 & \textbf{+0.9} & \textbf{+3.8} & +1.7 & +0.6 & +4.0 & +0.9 & +1.5 & 20.6 & 0.9$\times$ \\
 & DipMark & -1.1 & +2.3 & +1.5 & +0.3 & +0.4 & +1.4 & +0.8 & +1.2 & \textbf{+5.3} & +0.1 & +0.8 & 18.6 & 1.1$\times$ \\
 & KGW & \textbf{-17.4} & \textbf{+11.4} & \textbf{+4.2} & \textbf{+5.2} & \textbf{+1.9} & \textbf{+9.3} & \textbf{+7.0} & \textbf{+6.6} & \textbf{+20.2} & +0.3 & \textbf{+4.4} & 24.6 & \textbf{3.5$\times$} \\
 & PPL & \textbf{-7.7} & \textbf{+7.1} & \textbf{+6.7} & \textbf{+51.5} & \textbf{+2.1} & \textbf{+4.1} & \textbf{+3.9} & \textbf{+4.6} & \textbf{+13.0} & \textbf{+8.5} & +1.8 & 23.0 & 1.9$\times$ \\
 & SynthID & \textbf{-11.4} & \textbf{+24.0} & \textbf{+14.6} & \textbf{+25.9} & \textbf{+13.6} & \textbf{+12.3} & \textbf{+13.0} & \textbf{+7.9} & \textbf{+15.5} & \textbf{+13.8} & \textbf{+6.9} & 27.8 & \textbf{6.6$\times$} \\
\rowcolor{black!8} Ultramedical 70B & Unwatermarked & 86.6 & 11.0 & 11.8 & 1.0 & 0.0 & 8.0 & 0.5 & 1.5 & 23.2 & 2.2 & 2.4 & -- & 1.0$\times$ \\
 & AAR & -1.5 & +2.3 & +2.5 & +0.5 & +0.4 & \textbf{+4.5} & +0.9 & \textbf{+1.8} & \textbf{-13.1} & -0.9 & +0.6 & 10.7 & 1.0$\times$ \\
 & DipMark & +1.2 & +2.0 & +1.7 & 0.0 & 0.0 & +0.9 & +0.9 & +0.2 & -1.7 & +0.7 & +0.1 & 10.2 & 0.5$\times$ \\
 & KGW & -1.5 & +2.0 & +1.8 & +0.4 & +0.2 & +0.9 & +1.0 & +0.4 & \textbf{+10.3} & +1.4 & +0.6 & 10.3 & 3.0$\times$ \\
 & PPL & \textbf{-39.5} & \textbf{+6.5} & +0.2 & \textbf{+97.8} & \textbf{+1.3} & +2.7 & \textbf{+2.4} & \textbf{+7.4} & \textbf{+27.6} & \textbf{+22.3} & +2.6 & 11.4 & \textbf{22.0$\times$} \\
 & SynthID & \textbf{-9.2} & +3.9 & +3.3 & \textbf{+6.1} & +0.5 & \textbf{+3.5} & \textbf{+1.3} & \textbf{+2.7} & +4.1 & \textbf{+4.1} & +0.8 & 10.1 & 1.2$\times$ \\
\rowcolor{black!8} \textsc{Phi-4-14B} & Unwatermarked & 81.4 & 12.3 & 11.4 & 0.1 & 0.3 & 7.8 & 1.6 & 2.0 & 0.4 & 0.0 & 4.1 & -- & 1.0$\times$ \\
 & AAR & -2.5 & \textbf{+6.2} & \textbf{+6.5} & \textbf{+1.5} & \textbf{+1.8} & \textbf{+4.4} & \textbf{+2.9} & \textbf{+2.2} & +0.6 & +0.2 & \textbf{+4.0} & 11.3 & \textbf{14.0$\times$} \\
 & DipMark & -0.9 & \textbf{+3.9} & \textbf{+3.6} & +0.3 & +0.1 & +2.0 & +0.6 & -0.1 & +0.2 & 0.0 & +2.6 & 10.8 & 0.8$\times$ \\
 & KGW & -0.7 & \textbf{+9.2} & \textbf{+9.5} & \textbf{+2.2} & \textbf{+1.5} & \textbf{+5.5} & \textbf{+3.4} & \textbf{+3.6} & +0.2 & +0.1 & \textbf{+3.6} & 14.2 & \textbf{3.0$\times$} \\
 & PPL & \textbf{-8.6} & \textbf{+10.0} & \textbf{+9.6} & \textbf{+69.6} & \textbf{+3.8} & \textbf{+5.2} & \textbf{+3.2} & \textbf{+4.0} & \textbf{+5.4} & \textbf{+10.4} & +4.3 & 13.4 & \textbf{18.0$\times$} \\
 & SynthID & \textbf{-3.2} & \textbf{+13.7} & \textbf{+14.9} & \textbf{+10.6} & \textbf{+6.6} & \textbf{+9.4} & \textbf{+6.8} & \textbf{+5.5} & +0.7 & \textbf{+3.9} & \textbf{+8.0} & 15.4 & \textbf{7.6$\times$} \\
\rowcolor{black!8} \textsc{BioMistral-7B} & Unwatermarked & 40.8 & 1.1 & 1.3 & 0.3 & 0.4 & 1.8 & 1.9 & 0.7 & 7.4 & 0.0 & 0.4 & -- & 1.0$\times$ \\
 & AAR & -1.1 & \textbf{+21.2} & \textbf{+8.4} & \textbf{+5.2} & \textbf{+12.3} & \textbf{+8.6} & \textbf{+13.3} & \textbf{+8.2} & \textbf{+7.3} & \textbf{+3.4} & \textbf{+9.1} & 25.1 & \textbf{14.0$\times$} \\
 & DipMark & -1.3 & \textbf{+7.9} & \textbf{+3.4} & +0.9 & +1.4 & \textbf{+5.8} & \textbf{+3.8} & \textbf{+2.6} & \textbf{-6.5} & +0.2 & +0.9 & 11.5 & \textbf{4.9$\times$} \\
 & KGW & \textbf{-38.7} & +27.8 & -0.6 & \textbf{+19.6} & \textbf{+9.1} & \textbf{+16.9} & \textbf{+13.8} & \textbf{+11.7} & \textbf{+28.3} & \textbf{+3.9} & +16.7 & 48.9 & \textbf{18.1$\times$} \\
 & PPL & \textbf{-5.0} & \textbf{+13.2} & \textbf{+5.0} & \textbf{+18.5} & \textbf{+1.9} & \textbf{+8.6} & \textbf{+5.2} & \textbf{+5.3} & \textbf{+11.1} & \textbf{+2.0} & \textbf{+5.6} & 11.8 & \textbf{9.5$\times$} \\
 & SynthID & \textbf{-30.4} & \textbf{+10.5} & \textbf{+3.4} & \textbf{+82.3} & \textbf{+153.4} & +1.8 & \textbf{+30.4} & \textbf{+8.2} & \textbf{+66.7} & \textbf{+69.0} & +1.3 & 63.0 & \textbf{93.0$\times$} \\
\bottomrule
\end{tabular}
\begin{tablenotes}
\footnotesize
\item Single-response axes (per $100$ questions where suffixed): ConfErr (confident error, on BOTH-CORRECT pairs), F\&C (faulty-but-correct: correct letter $+\geq 1$ defect), Spell/100, Fab/100 (fabricated entities), Contr (within-response contradiction), VigFab (vignette fabrication), Misread (clue misread), Fmt (format/termination failure), Ling (linguistic corruption), Misapp/100 (misapplied real terms). Pairwise: Major\% (MAJOR divergence on BOTH-CORRECT pairs), FabRatio (WM/Base vignette-fabrication rate).
\end{tablenotes}
\end{threeparttable}
\end{sidewaystable}

\begin{sidewaystable}
\centering
\scriptsize
\setlength{\tabcolsep}{1.5pt}
\begin{threeparttable}
\caption{\textbf{Full reasoning audit at $99\%$ TPR on MedXpertQA-MM}, for all $7$ VLMs.
We report the accuracy ($\Delta$Acc), the confident-error rate ($\Delta$ConfErr), the faulty-but-correct rate ($\Delta$F\&C), the percentage of corrupted spellings ($\Delta$Spell/100), the percentage of fabricated facts ($\Delta$Fab/100), the image-grounding net change ($\Delta$Net\%), the within-response contradiction rate ($\Delta$Contr), vignette fabrication ($\Delta$VigFab), clue misreads ($\Delta$Misread), formatting failures ($\Delta$Fmt), linguistic corruption ($\Delta$Ling), the percentage of misapplied medical terms ($\Delta$Misapp/100), supported visual claims ($\Delta$Sup), contradicted visual claims ($\Delta$Con), the percentage of paired responses with major diagnostic divergence (Major\%), and the vignette fabrication ratio (FabRatio).
For unwatermarked rows we report the actual value, whereas for watermarked rows we report the difference/ratio with the unwatermarked baseline.
We \textbf{bold} entries that are statistically significant (p-value below $0.01$).
}\label{tab:taskA_full_MedXpertQA_MM}
\begin{tabular}{llrrrrrrrrrrrrrrrr}
\toprule
Model & Scheme & $\Delta$Acc & $\Delta$ConfErr & $\Delta$F\&C & $\Delta$Spell/100 & $\Delta$Fab/100 & $\Delta$Net\% & $\Delta$Contr & $\Delta$VigFab & $\Delta$Misread & $\Delta$Fmt & $\Delta$Ling & $\Delta$Misapp/100 & $\Delta$Sup & $\Delta$Con & Major\% & FabRatio \\
\midrule
\rowcolor{black!8} \textsc{Gemma-4-31B} & Unwatermarked & 56.2 & 12.2 & 9.2 & 0.1 & 0.3 & 2 & 5.0 & 0.8 & 1.9 & 0.1 & 0.0 & 1.7 & 2.55 & 0.53 & -- & 1.0$\times$ \\
 & AAR & -0.6 & -0.6 & -0.3 & +0.9 & +0.3 & +1 & +0.7 & +0.1 & -0.1 & +0.1 & +0.8 & -1.1 & -0.01 & -0.03 & 6.0 & 3.0$\times$ \\
 & DipMark & -0.1 & -1.0 & -0.3 & +0.5 & +0.2 & +2 & +0.5 & -0.5 & 0.0 & +0.1 & 0.0 & -0.2 & +0.02 & -0.03 & 3.6 & 2.0$\times$ \\
 & KGW & +0.2 & -1.7 & -0.3 & \textbf{+1.8} & +0.2 & +1 & +1.7 & -0.2 & +0.3 & +0.2 & \textbf{+1.5} & +0.2 & +0.04 & +0.01 & 6.8 & 0.5$\times$ \\
 & PPL & -0.9 & +1.6 & +3.1 & \textbf{+34.5} & \textbf{+1.5} & -2 & +1.8 & -0.3 & \textbf{+3.4} & \textbf{+2.1} & \textbf{+22.5} & 0.0 & -0.04 & -0.01 & 5.0 & \textbf{6.0$\times$} \\
 & SynthID & -1.0 & +0.4 & -0.3 & \textbf{+7.6} & +0.8 & +3 & +1.0 & -0.1 & -0.3 & +0.1 & \textbf{+11.0} & -0.9 & +0.02 & -0.04 & 6.8 & 4.0$\times$ \\
\rowcolor{black!8} \textsc{Lingshu-7B} & Unwatermarked & 23.9 & 55.3 & 13.7 & 0.4 & 1.5 & 0 & 17.0 & 3.5 & 7.6 & 0.8 & 0.5 & 12.1 & 1.15 & 0.77 & -- & 1.0$\times$ \\
 & AAR & -0.5 & +13.8 & +3.7 & \textbf{+5.4} & \textbf{+12.9} & \textbf{-67} & +3.5 & \textbf{+7.9} & \textbf{+3.8} & \textbf{+2.0} & \textbf{+3.5} & +18.1 & \textbf{-0.15} & +0.09 & 33.0 & \textbf{2.8$\times$} \\
 & DipMark & +1.0 & +5.4 & +1.4 & +0.1 & +0.3 & -6 & +2.7 & +0.1 & 0.0 & -0.1 & +0.1 & +2.7 & -0.02 & -0.00 & 22.6 & 1.0$\times$ \\
 & KGW & -0.8 & \textbf{+28.6} & +6.0 & \textbf{+16.2} & \textbf{+13.2} & \textbf{-103} & \textbf{+20.2} & \textbf{+10.4} & \textbf{+11.0} & \textbf{+5.4} & \textbf{+4.6} & +2.4 & -0.12 & \textbf{+0.37} & 42.9 & \textbf{3.6$\times$} \\
 & PPL & -0.6 & +7.3 & +2.3 & \textbf{+37.9} & \textbf{+3.9} & -44 & +3.9 & \textbf{+3.0} & \textbf{+7.9} & \textbf{+5.1} & \textbf{+8.4} & +11.5 & \textbf{-0.13} & +0.03 & 20.0 & \textbf{2.4$\times$} \\
 & SynthID & -2.1 & \textbf{+26.2} & \textbf{+5.5} & \textbf{+18.9} & \textbf{+39.2} & \textbf{-105} & \textbf{+13.2} & \textbf{+16.8} & \textbf{+8.5} & \textbf{+7.7} & \textbf{+13.5} & \textbf{+20.2} & \textbf{-0.14} & \textbf{+0.28} & 30.5 & \textbf{7.0$\times$} \\
\rowcolor{black!8} \textsc{Lingshu-32B} & Unwatermarked & 31.1 & 40.8 & 15.1 & 0.8 & 0.1 & 1 & 9.6 & 2.0 & 3.7 & 0.4 & 0.1 & 5.9 & 1.38 & 0.68 & -- & 1.0$\times$ \\
 & AAR & -1.8 & +9.2 & +1.3 & \textbf{+2.9} & \textbf{+5.4} & \textbf{-46} & \textbf{+4.8} & \textbf{+5.5} & \textbf{+3.4} & \textbf{+2.5} & \textbf{+3.5} & +2.0 & \textbf{-0.15} & \textbf{+0.17} & 18.0 & \textbf{3.8$\times$} \\
 & DipMark & +0.4 & -1.2 & -1.0 & -0.1 & +0.4 & -16 & +1.0 & +0.1 & +0.6 & -0.2 & 0.0 & 0.0 & -0.05 & +0.07 & 15.3 & 0.8$\times$ \\
 & KGW & -4.0 & +10.1 & -0.1 & \textbf{+1.9} & \textbf{+1.8} & \textbf{-39} & \textbf{+8.9} & \textbf{+2.4} & \textbf{+4.8} & +1.1 & +0.8 & +3.6 & -0.05 & \textbf{+0.23} & 14.4 & 1.8$\times$ \\
 & PPL & -2.4 & -3.2 & -2.0 & \textbf{+27.4} & +0.8 & -26 & +2.3 & +1.2 & \textbf{+7.1} & \textbf{+6.7} & \textbf{+5.3} & -0.6 & -0.06 & \textbf{+0.12} & 11.2 & \textbf{2.2$\times$} \\
 & SynthID & -3.8 & +6.7 & +2.0 & \textbf{+23.9} & \textbf{+13.7} & \textbf{-56} & \textbf{+16.2} & \textbf{+9.8} & \textbf{+9.5} & \textbf{+8.8} & \textbf{+13.2} & +5.2 & -0.12 & \textbf{+0.28} & 21.7 & \textbf{4.9$\times$} \\
\rowcolor{black!8} \textsc{MedGemma-4B} & Unwatermarked & 24.1 & 56.9 & 16.5 & 1.0 & 1.9 & 0 & 29.7 & 3.5 & 9.0 & 7.7 & 0.0 & 11.1 & 1.24 & 1.11 & -- & 1.0$\times$ \\
 & AAR & -1.0 & \textbf{+13.0} & +1.2 & \textbf{+3.4} & \textbf{+6.5} & -108 & +2.1 & \textbf{+7.4} & +3.3 & \textbf{+11.2} & \textbf{+4.6} & +8.1 & \textbf{+0.15} & \textbf{+0.29} & 38.3 & \textbf{3.1$\times$} \\
 & DipMark & +0.2 & +11.6 & +1.6 & +0.2 & +0.7 & -103 & -0.9 & +0.1 & -0.2 & -0.2 & +0.3 & +2.0 & -0.08 & +0.06 & 35.0 & 1.2$\times$ \\
 & KGW & -2.6 & \textbf{+19.2} & +0.6 & \textbf{+15.4} & \textbf{+8.9} & \textbf{-266} & \textbf{+11.2} & \textbf{+6.5} & \textbf{+7.4} & \textbf{+13.1} & \textbf{+3.5} & +16.2 & -0.01 & \textbf{+0.32} & 31.7 & \textbf{3.1$\times$} \\
 & PPL & -1.7 & +8.3 & 0.0 & \textbf{+43.6} & \textbf{+6.0} & -46 & +4.2 & \textbf{+3.1} & \textbf{+8.2} & \textbf{+8.0} & \textbf{+9.7} & +5.0 & -0.01 & +0.05 & 34.7 & \textbf{2.1$\times$} \\
 & SynthID & -3.1 & \textbf{+24.3} & +0.4 & \textbf{+18.0} & \textbf{+12.6} & \textbf{-168} & \textbf{+8.5} & \textbf{+11.1} & \textbf{+8.1} & \textbf{+22.1} & \textbf{+8.7} & \textbf{+24.3} & -0.05 & \textbf{+0.20} & 36.0 & \textbf{4.2$\times$} \\
\rowcolor{black!8} \textsc{MedGemma-27B} & Unwatermarked & 33.7 & 41.5 & 16.9 & 0.6 & 2.3 & 0 & 12.4 & 1.6 & 5.8 & 2.8 & 0.0 & 9.1 & 1.90 & 1.59 & -- & 1.0$\times$ \\
 & AAR & -0.9 & -3.0 & -0.8 & \textbf{+1.5} & +1.9 & -43 & +2.6 & \textbf{+2.9} & -0.7 & \textbf{+2.6} & +0.1 & -2.0 & -0.08 & +0.06 & 19.5 & 1.6$\times$ \\
 & DipMark & +1.2 & -5.8 & -0.3 & +0.3 & -0.2 & -16 & +0.2 & +0.1 & -1.1 & +0.7 & +0.1 & -1.0 & -0.06 & +0.00 & 18.4 & 1.0$\times$ \\
 & KGW & -1.0 & +4.7 & +1.9 & +0.7 & +0.3 & -41 & +3.3 & +1.6 & +0.8 & +1.6 & +0.2 & +3.1 & -0.01 & +0.13 & 18.5 & \textbf{1.8$\times$} \\
 & PPL & -0.2 & +7.4 & +3.4 & \textbf{+34.2} & \textbf{+2.7} & -23 & \textbf{+5.5} & +1.5 & \textbf{+4.5} & +1.8 & \textbf{+6.7} & -3.0 & -0.04 & +0.04 & 23.0 & 1.4$\times$ \\
 & SynthID & -0.3 & -1.9 & -0.8 & +1.5 & +1.3 & +3 & +2.7 & +0.9 & -0.3 & +0.3 & +0.2 & -1.9 & +0.01 & +0.00 & 22.8 & 0.9$\times$ \\
\rowcolor{black!8} \textsc{Qwen-3-VL-8B} & Unwatermarked & 31.3 & 47.0 & 19.9 & 0.6 & 1.8 & 0 & 21.7 & 3.0 & 4.9 & 7.6 & 0.6 & 11.0 & 1.95 & 1.57 & -- & 1.0$\times$ \\
 & AAR & \textbf{-4.3} & 0.0 & \textbf{-5.2} & 0.0 & \textbf{+2.5} & -3 & +1.3 & +0.6 & +0.7 & \textbf{+7.3} & +0.6 & +7.3 & +0.06 & +0.07 & 22.2 & 1.8$\times$ \\
 & DipMark & -2.5 & +2.9 & -1.5 & -0.1 & +1.8 & +17 & +1.5 & +1.1 & +2.6 & +1.2 & -0.2 & +2.3 & +0.06 & -0.01 & 25.0 & 1.1$\times$ \\
 & KGW & -1.0 & +1.1 & +0.7 & +0.7 & +1.9 & +8 & +4.5 & +2.2 & +1.7 & +2.9 & +0.1 & +3.4 & +0.07 & +0.04 & 24.6 & 1.7$\times$ \\
 & PPL & \textbf{-5.1} & +2.0 & -2.5 & \textbf{+22.5} & +1.1 & -22 & -1.1 & +0.6 & \textbf{+4.7} & \textbf{+7.3} & \textbf{+7.2} & +4.7 & -0.10 & -0.02 & 24.9 & 1.5$\times$ \\
 & SynthID & \textbf{-4.5} & +3.1 & -2.1 & +0.1 & +1.4 & -9 & +2.0 & +1.9 & +0.6 & \textbf{+6.2} & -0.4 & +5.0 & -0.04 & -0.00 & 24.5 & 1.0$\times$ \\
\rowcolor{black!8} \textsc{Qwen-3-VL-32B} & Unwatermarked & 40.8 & 27.9 & 15.7 & 0.2 & 0.7 & 1 & 15.3 & 1.3 & 3.8 & 3.8 & 0.2 & 3.3 & 2.69 & 1.29 & -- & 1.0$\times$ \\
 & AAR & -3.0 & +3.1 & -0.7 & \textbf{+1.5} & \textbf{+1.9} & -2 & -1.1 & +0.3 & -0.1 & +0.5 & \textbf{+3.0} & +2.7 & -0.01 & +0.02 & 15.4 & 0.6$\times$ \\
 & DipMark & -1.4 & +5.1 & +1.0 & +0.5 & +0.9 & -3 & -1.0 & +0.1 & 0.0 & +0.6 & +0.8 & +4.0 & -0.00 & +0.04 & 14.1 & 0.8$\times$ \\
 & KGW & -0.9 & +3.9 & -0.1 & +1.1 & +0.2 & +3 & +1.4 & +0.8 & -0.2 & +0.6 & \textbf{+1.2} & +2.1 & +0.08 & +0.04 & 15.4 & 1.0$\times$ \\
 & PPL & -2.6 & +4.1 & -0.6 & \textbf{+35.6} & \textbf{+2.0} & -3 & +1.2 & -0.5 & \textbf{+2.6} & +2.2 & \textbf{+11.6} & +1.6 & +0.01 & +0.05 & 14.4 & 1.2$\times$ \\
 & SynthID & +1.1 & +3.9 & +2.5 & +1.2 & \textbf{+2.3} & -9 & -0.4 & +0.9 & -0.4 & +0.7 & \textbf{+3.5} & +3.9 & -0.04 & +0.08 & 15.3 & 1.6$\times$ \\
\bottomrule
\end{tabular}
\begin{tablenotes}
\footnotesize
\item Single-response axes (per $100$ questions where suffixed): ConfErr (confident error, on BOTH-CORRECT pairs), F\&C (faulty-but-correct: correct letter $+\geq 1$ defect), Spell/100, Fab/100 (fabricated entities), Contr (within-response contradiction), VigFab (vignette fabrication), Misread (clue misread), Fmt (format/termination failure), Ling (linguistic corruption), Misapp/100 (misapplied real terms). Image-grounding (multimodal): Sup, Con (per-question \textsc{supported}/\textsc{contradicted} visual claims), Net\% (relative change in $\textsc{supported}-\textsc{contradicted}$). Pairwise: Major\% (MAJOR divergence on BOTH-CORRECT pairs), FabRatio (WM/Base vignette-fabrication rate).
\end{tablenotes}
\end{threeparttable}
\end{sidewaystable}

\begin{figure}[t]
    \centering
    \includegraphics[width=\textwidth]{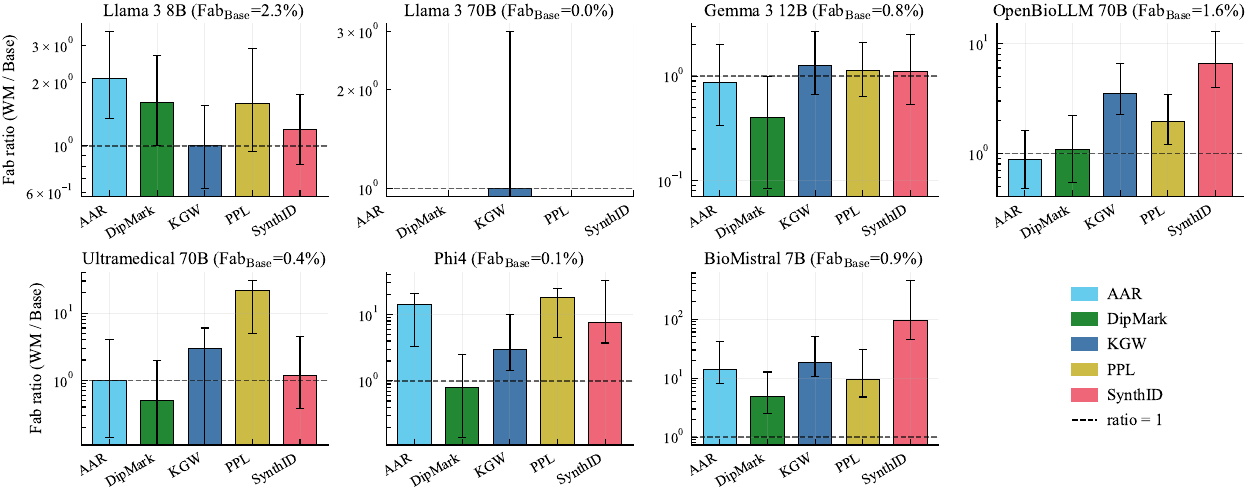}
    \caption{\textbf{Pairwise vignette-fabrication ratio per scheme (MedQA, log scale).} WM/Base rate ratio per (model, scheme) cell; dashed line at $1$ (no effect). Cells with very low baseline rates have wide CIs.}
    \label{fig:taskC_fabratio_medqa}
\end{figure}

\begin{figure}[t]
    \centering
    \includegraphics[width=\textwidth]{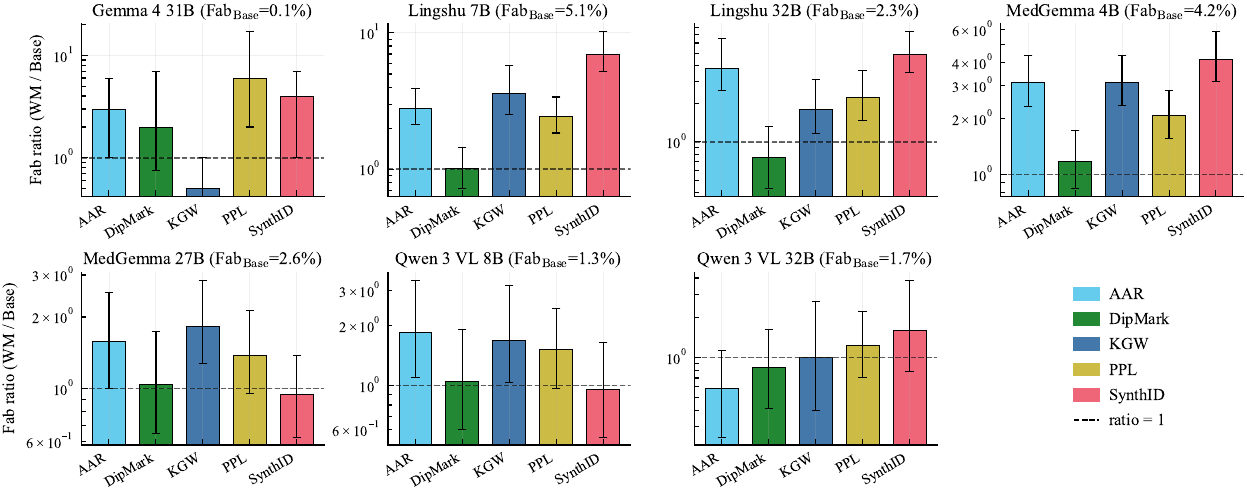}
    \caption{\textbf{Pairwise vignette-fabrication ratio per scheme (MedXpertQA-MM, log scale).} Conventions as in \cref{fig:taskC_fabratio_medqa}.}
    \label{fig:taskC_fabratio_mm}
\end{figure}

\begin{figure}[t]
    \centering
    \includegraphics[width=\textwidth]{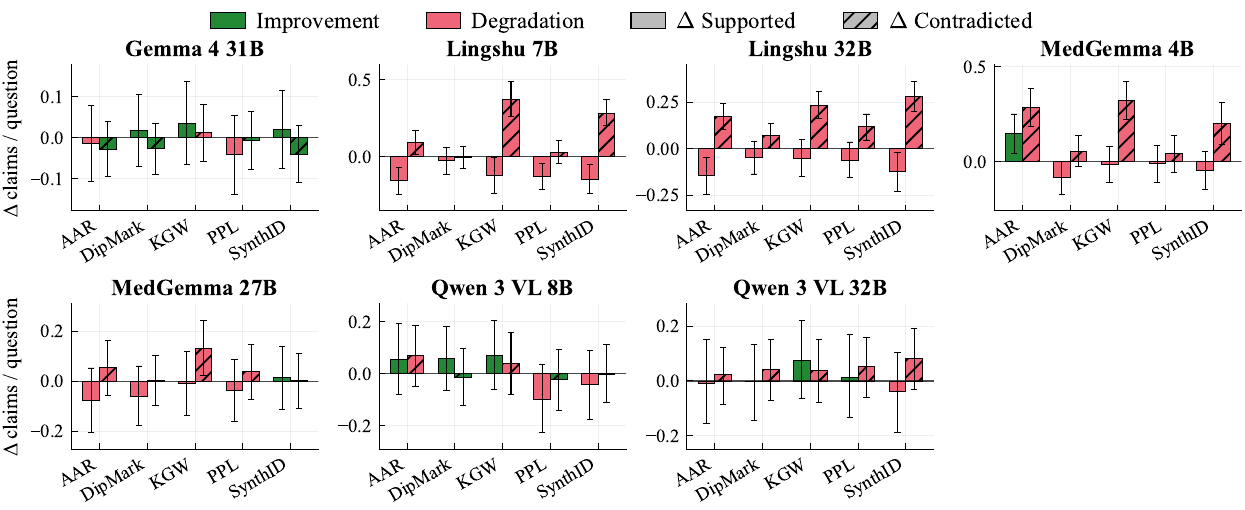}
    \caption{\textbf{Per-scheme $\Delta$ in image-supported and image-contradicted visual claims, all $7$ VLMs (MedXpertQA-MM).} All-VLM companion to \cref{fig:image_decomp}. Within each panel the left bar is $\Delta$ Supported (solid) and the right bar is $\Delta$ Contradicted (hatched). Bars are coloured green where the change improves visual grounding ($\Delta$ Supported~$>0$ or $\Delta$ Contradicted~$<0$) and red otherwise.}
    \label{fig:image_decomp_full}
\end{figure}

\newpage

\subsection{Reasoning Models: Per-Dimension and Length Detail}
\label{app:thinking_detail}

\paragraph{Per-dimension judge scores (\cref{sec:thinking}).}
\cref{fig:thinking_dim_scores} shows the mean pairwise
Full~WM--reference score on all six judge dimensions, with the two
reference conditions (vs.\ Base, vs.\ FA-only) plotted side by side.
The two series overlap within their 95\% bootstrap intervals on every
panel: this is equivalent to FA-only $\approx$ Base on every axis,
since (Full WM $-$ Base) $-$ (Full WM $-$ FA-only) $=$ FA-only $-$
Base.
Across all four models the pattern at $\delta = 5$ is the same:
\emph{reasoning efficiency} and \emph{terminological precision} drop
substantially, \emph{clinical coherence} and \emph{distractor
engagement} drop more mildly, and \emph{answer--reasoning alignment}
and \emph{diagnostic accuracy} stay near zero.

\begin{figure}[h]
  \centering
  \includegraphics[width=\linewidth]{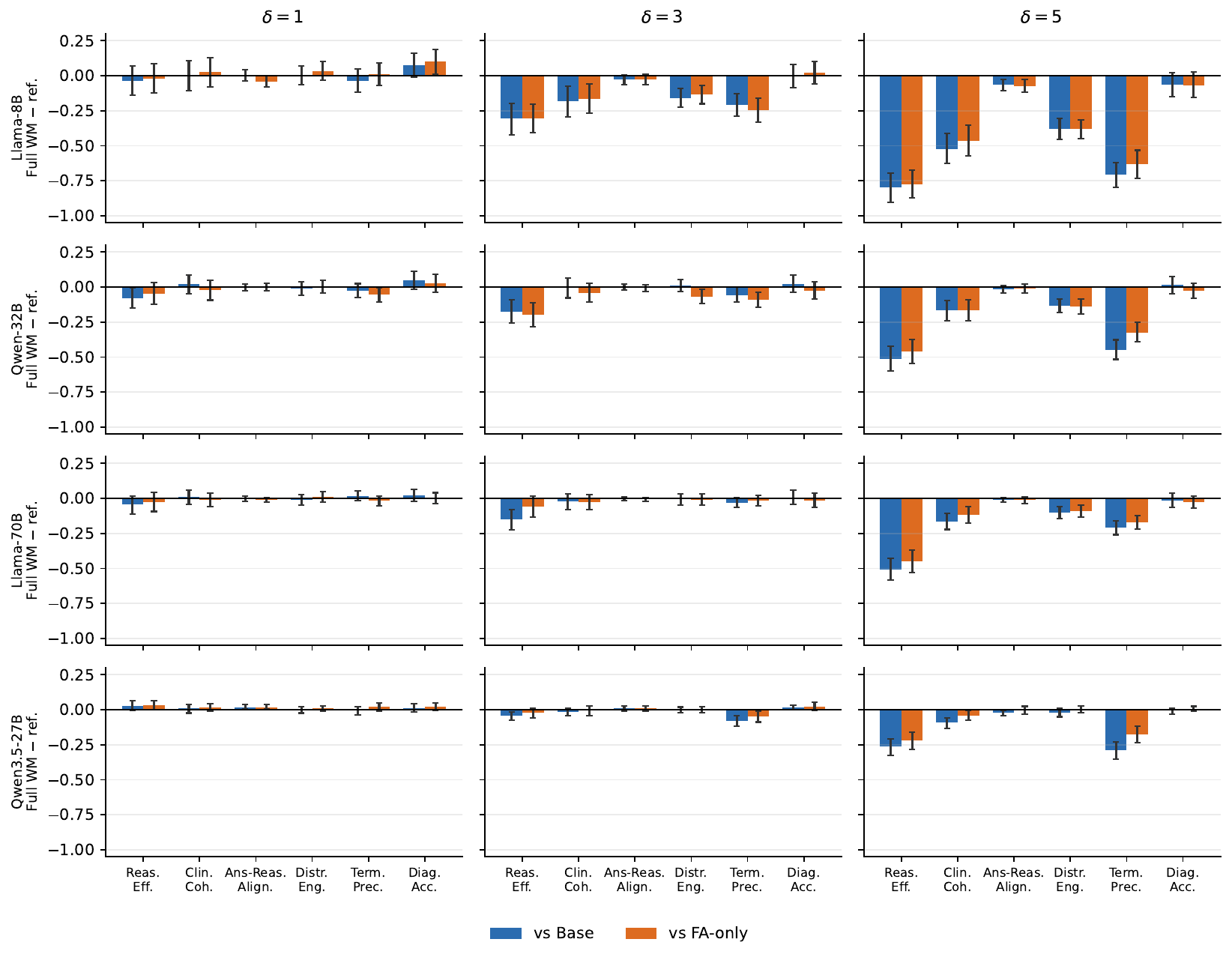}
  \caption{\textbf{Pairwise mean Full~WM--reference score on six
    reasoning-quality axes}, four reasoning models $\times$ three KGW
    strengths. Blue: Full~WM scored against the unwatermarked Base
    response; orange: Full~WM scored against the FA-only response.
    Per-question
    scores are integers in $\{-2,-1,0,+1,+2\}$ (rubric in
    \cref{app:audit3_prompt}); panel values are sample means.
    Negative bar = Full~WM judged worse than the reference. Error
    bars: 95\% bootstrap intervals.}
  \label{fig:thinking_dim_scores}
\end{figure}

\paragraph{Clinical-deployment ratings.}
\cref{tab:thinking_safety} gives the absolute
\textsc{harm\_risk\_present} and \textsc{major\_revision} rates per
condition.
Only \textsc{Llama-3.1-8B} exhibits a Full~WM-specific safety jump
($+12.8$\,pp HARM at $\delta=5$); on the three larger models the
Full~WM HARM\% sits within $\pm 2$\,pp of the No~WM baseline at every
$\delta$.

\begin{table}[h]
\centering
\footnotesize
\caption{HARM\_RISK\_PRESENT and MAJOR\_REVISION rates by watermarking scope (KGW). No~WM is judged in pairwise mode against Full~WM (small contrast bias) and is delta-independent; FA-only and Full~WM are absolute per-response ratings.}
\label{tab:thinking_safety}
\begin{tabular}{ll rr rr rr}
\toprule
 &  & \multicolumn{2}{c}{No WM} & \multicolumn{2}{c}{FA-only} & \multicolumn{2}{c}{Full WM} \\
\cmidrule(lr){3-4}\cmidrule(lr){5-6}\cmidrule(lr){7-8}
Model & $\delta$ & HARM\% & MAJOR\% & HARM\% & MAJOR\% & HARM\% & MAJOR\% \\
\midrule
  \multirow{3}{*}{\textsc{Llama-8B}} & 1 & \multirow{3}{*}{50.9} & \multirow{3}{*}{48.8} & 56.6 & 56.2 & 58.7 & 58.1 \\
   & 3 &  &  & 58.2 & 57.6 & 57.1 & 56.7 \\
   & 5 &  &  & 55.7 & 55.3 & 63.7 & 63.1 \\
\midrule
  \multirow{3}{*}{\textsc{Qwen-32B}} & 1 & \multirow{3}{*}{22.1} & \multirow{3}{*}{20.7} & 16.2 & 16.0 & 16.2 & 16.2 \\
   & 3 &  &  & 18.8 & 18.8 & 17.5 & 17.5 \\
   & 5 &  &  & 18.9 & 18.6 & 19.7 & 19.5 \\
\midrule
  \multirow{3}{*}{\textsc{Llama-70B}} & 1 & \multirow{3}{*}{10.0} & \multirow{3}{*}{9.3} & 8.0 & 8.0 & 10.0 & 10.0 \\
   & 3 &  &  & 8.0 & 8.0 & 11.0 & 11.0 \\
   & 5 &  &  & 8.8 & 8.8 & 9.0 & 9.0 \\
\midrule
  \multirow{3}{*}{\textsc{Qwen-3.5-27B}} & 1 & \multirow{3}{*}{3.7} & \multirow{3}{*}{3.7} & 2.8 & 2.8 & 2.6 & 2.6 \\
   & 3 &  &  & 2.4 & 2.4 & 1.6 & 1.6 \\
   & 5 &  &  & 2.1 & 2.5 & 2.6 & 2.6 \\
\bottomrule
\end{tabular}
\end{table}

\clearpage
\section{Judge Validation}
\label{app:judge_validation}

\subsection{Inter-Rater Agreement}
\label{app:judge_agreement_full}

\cref{tab:judge_agreement_full} reports per-axis agreement between the \textsc{Gemini-3-Flash} judge, the physician annotator, and \textsc{Claude-Sonnet-4.6} on the full $650$ blinded samples.
The main-text summary (\cref{sec:human_validation}) restricts to the four axes underlying the headline claims; the table below provides the complete picture, including low-prevalence stylistic flags where agreement is expectedly lower.

\begin{table}[h]
\centering
\small
\caption{\textbf{Full judge--physician and judge--judge agreement} on $650$ blinded samples. $\kappa$: Cohen's kappa; $\rho$: Spearman correlation on per-response counts; $\kappa_w$: linearly weighted kappa.}
\label{tab:judge_agreement_full}
\begin{tabular}{lcc}
\toprule
Axis & \makecell{\textsc{Gemini-3-Flash}\\vs.\ Human-expert} & \makecell{\textsc{Gemini-3-Flash}\\vs.\ \textsc{Claude-Sonnet-4.6}} \\
\midrule
\multicolumn{3}{l}{\textit{Single-response binary flags ($n{=}500$)}} \\
Vignette fabrication       & 0.61 & 0.62 \\
Confident error            & 0.76 & 0.68 \\
Self-contradiction         & 0.60 & 0.45 \\
Clue misread               & 0.44 & 0.48 \\
Format / termination       & 0.66 & 0.39 \\
Linguistic corruption      & 0.73 & 0.44 \\
\midrule
\multicolumn{3}{l}{\textit{Counts ($n{=}500$; $\kappa$/binarised, $\rho$/raw)}} \\
Fabricated entities        & 0.75 / 0.75 & 0.80 / 0.81 \\
Misapplied terms           & 0.42 / 0.47 & 0.50 / 0.52 \\
Corrupted spellings        & 0.71 / 0.71 & 0.50 / 0.51 \\
\midrule
\multicolumn{3}{l}{\textit{Image grounding ($n{=}250$, MM only)}} \\
Contradicted claims ($\rho$) & 0.74 & 0.50 \\
\midrule
\multicolumn{3}{l}{\textit{Pairwise divergence ($n{=}150$)}} \\
3-level ($\kappa_w$)       & 0.65 & 0.66 \\
Major vs.\ not ($\kappa$)  & 0.67 & 0.72 \\
\bottomrule
\end{tabular}
\end{table}

\subsection{Clinician Validation Panel}
\label{app:clinician_validation}

\paragraph{Purpose and scope.}
Each watermarked configuration in this panel aggregates only $5$ scored outputs, so the analysis is a qualitative end-to-end check, not a statistically powered validation.
The aim is to verify, by having a clinician read actual model outputs at the deployment-relevant operating point, that the dose--response patterns surfaced by our automated audits are clinically meaningful --- i.e.\ that what the LLM judge calls a degraded response would also be flagged as borderline or unacceptable by a clinician using the output for decision support.
This panel is distinct from the per-axis inter-rater agreement reported in \cref{sec:human_validation}.

\paragraph{Models.}
We deliberately picked two contrasting MedXpertQA-MM models: \emph{\textsc{Lingshu-7B}} is a recent medical-specialised VLM that was among the
 leaders on the original MedXpertQA-MM benchmark in its size class~\citep{xu2025lingshu}, while \emph{\textsc{Gemma-4-31B}} is the most
  recent frontier general-purpose VLM in our pool.

\paragraph{Rating protocol and analysis.}
Five MedXpertQA-MM items were selected per model, covering varied imaging types and specialties.
 For each item, we generated $50$ unwatermarked completions (different sampling seeds for tighter baseline variance estimation) and one watermarked completion
across 12 different levels of detectability. A human expert scored the answers on a simple, 3-level rubric, with the answers in randomized order: \emph{Accept}
(correct letter, or a defensible alternative, with sound medical reasoning), \emph{Borderline} (correct letter with a flawed reasoning chain, or a
defensible alternative letter), or \emph{Reject} (wrong letter without a defensible explanation, hallucinated findings, or self-contradictory reasoning).
The harm rate is computed as $(n_{\text{reject}}
   + 0.5 \cdot n_{\text{borderline}})/n$. To anchor each model's intrinsic noise floor, the unwatermarked $250$-output pool ($5$ items $\times$ $50$ seeds) yields
    a $95\%$ binomial confidence band on the unwatermarked harm rate; we treat this band as the model's ``noise channel'' and call any (scheme, strength)
    configuration whose dot lies outside this band a clinician-detectable degradation.

\paragraph{Results.}
\cref{fig:clinician_dose_response} shows clear patterns at the deployment-level operating point (TPR~$>90\%$).

\begin{figure}[t]
    \centering
    \includegraphics[width=\linewidth]{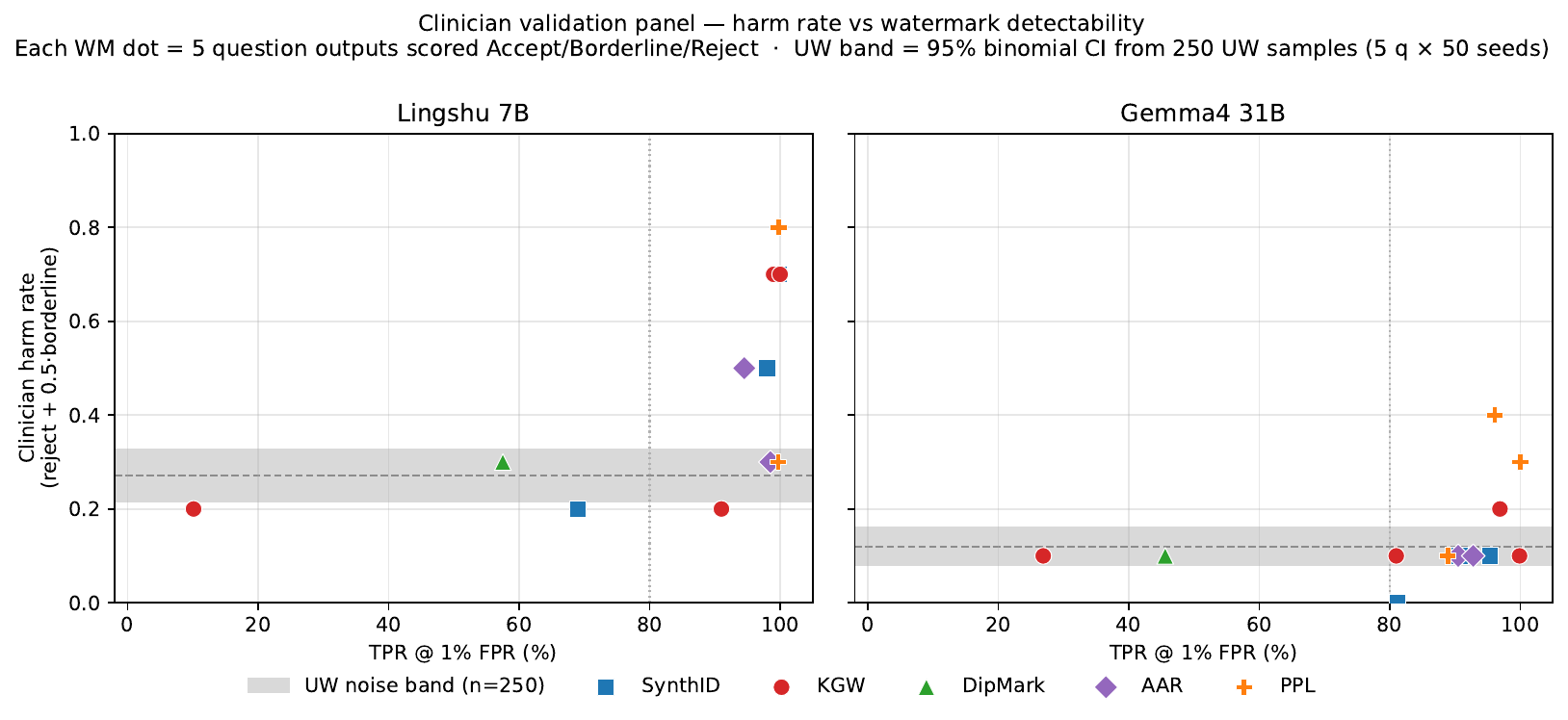}
    \caption{Clinician validation panel -- harm rate vs.\ watermark detectability. Each watermarked point aggregates $5$ outputs (one per item) at a fixed (scheme, strength); the unwatermarked noise band is the $95\%$ binomial CI from the $250$-sample unwatermarked pool ($5$ items $\times$ $50$ seeds). Left: \textsc{Lingshu-7B} (medical-specialised). Right: \textsc{Gemma-4-31B} (general-purpose frontier). Dotted vertical guide marks TPR~$=$~$80\%$. At the deployment-level operating point (TPR~$\approx$~$99\%$), Lingshu's strongest watermark configurations land at clinician harm rates $0.5$--$0.8$, well outside its $\sim$$0.27$ unwatermarked band; Gemma's land at or just inside its $\sim$$0.12$ unwatermarked band, with only $\mathrm{PPL}$ at top strength visibly escaping.}
    \label{fig:clinician_dose_response}
\end{figure}

\emph{\textsc{Lingshu-7B} is highly vulnerable.} The unwatermarked noise band sits around a harm rate of $0.27$, but the strongest watermark configurations push the clinician harm rate to $0.5$--$0.8$ ($\mathrm{PPL}$, the strongest $\mathrm{KGW}$ setting, $\mathrm{AAR}$, and $\mathrm{SynthID}$ at high strength), well outside the noise channel and corresponding to ratings dominated by \emph{Reject}. Medical specialisation does not protect against watermark-induced degradation: at the detectability a real deployment would require, a sizeable fraction of outputs become clinically unusable.

\emph{\textsc{Gemma-4-31B} remains within its noise channel.} The unwatermarked harm rate sits around $0.12$, and at TPR~$\geq$~$95\%$ most schemes (KGW, DipMark, AAR, SynthID) produce dots inside or directly adjacent to the unwatermarked band. Only $\mathrm{PPL}$ at the strongest setting visibly escapes ($\approx$~$0.40$). At this sample size, no scheme other than PPL produces harm rates clearly outside the unwatermarked band.

This two-model split confirms the pattern observed in the automated reasoning-quality audits (\cref{sec:results}): watermark vulnerability is model-specific and not predicted by benchmark accuracy or domain specialisation alone.

\clearpage
\section{Branching-Point Experiment}
\label{app:branching}

The reasoning failures documented in \cref{sec:reasoning_degradation}
are heterogeneous---fabricated entities, misapplied terminology,
contradicted visual claims---but they share a suggestive pattern:
the model commits to the wrong clinical interpretation early in its
response, and the rest of the reasoning chain follows from that
commitment.
We hypothesise that watermarking is particularly harmful at
\emph{branching points}: tokens where the model decides between
competing clinical interpretations of the same evidence.
At such tokens, even a small shift in the sampling distribution can
tip the model toward a wrong interpretation that it would otherwise
dismiss, locking in an error that propagates through the entire
response.
To test this, we design a controlled experiment that seeds the model
with a short clinical prefix and then lets the watermark act at the
immediately following decision token.
If the hypothesis is correct, watermarking should amplify misleading
prefixes (making the model follow a wrong cue it would normally
override) and suppress helpful ones (preventing the model from
exploiting a correct cue it would normally use).

\paragraph{Protocol.}
We construct branching variants of three MedXpertQA-MM
questions\footnote{%
  Q1: MEN2B / medullary thyroid carcinoma;
  Q2: post-streptococcal glomerulonephritis (PSGN);
  Q3: peripheral blood smear (basophilia).}
by seeding the model's response with a short clinical prefix so that
the watermark first acts at the immediately following decision token.
\emph{Rescue} prefixes describe evidence pointing toward the correct
answer; \emph{distractor} prefixes describe evidence consistent with a
specific wrong interpretation.
Both can be \emph{perceptual} (purportedly describing what the image
shows) or \emph{inferential} (framing a clinical interpretation or
reasoning step).
Constructing clinically plausible prefixes requires domain expertise,
which limits the experiment to a small number of carefully selected
questions spanning distinct reasoning types (visual interpretation,
evidence integration, differential diagnosis).
For each (question, prefix, scheme) configuration we draw $N{=}200$
paired trials in which we also vary the watermark private key.
Comparisons are paired by seed against the unwatermarked condition.
All tests use \textsc{Qwen-3-VL-32B} at temperature~$0.7$.

\begin{figure}[!htbp]
\centering

\definecolor{qboxbg}{RGB}{248,248,245}
\definecolor{qboxrule}{RGB}{120,120,120}
\definecolor{prefixbg}{RGB}{255,247,224}
\definecolor{prefixrule}{RGB}{180,140,40}

{\setlength{\fboxsep}{8pt}%
\fcolorbox{qboxrule}{qboxbg}{%
\begin{minipage}{0.96\linewidth}
\footnotesize
\textbf{Question 2 (MedXpertQA-MM 503).}\quad
A 5-year-old boy is brought to the pediatrician with progressive
swelling that began in the face and spread to the extremities over
the past 24 hours. Medical history is unremarkable except for a
honey-coloured rash that resolved two weeks ago. Urinalysis reveals
hematuria with red blood cell casts. A renal biopsy is performed.
\emph{Which of the following light-microscopy findings would most
likely be observed?}

\smallskip
\textcolor{green!45!black}{\textbf{(A) Figure A \cmark}}\enspace
\textbf{(B)}\,Figure\,B\enspace
\textbf{(C)}\,Figure\,D\enspace
\textbf{(D)}\,Figure\,C\enspace
\textbf{(E)}\,Figure\,E\enspace
\textit{(post-streptococcal glomerulonephritis).}

\smallskip
\centering
\begin{tabular}{@{}ccccc@{}}
\includegraphics[width=0.165\linewidth,height=0.12\linewidth,keepaspectratio=false]{figures/branching_images/MM-1389-a.jpeg} &
\includegraphics[width=0.165\linewidth,height=0.12\linewidth,keepaspectratio=false]{figures/branching_images/MM-1389-b.jpeg} &
\includegraphics[width=0.165\linewidth,height=0.12\linewidth,keepaspectratio=false]{figures/branching_images/MM-1389-c.jpeg} &
\includegraphics[width=0.165\linewidth,height=0.12\linewidth,keepaspectratio=false]{figures/branching_images/MM-1389-d.jpeg} &
\includegraphics[width=0.165\linewidth,height=0.12\linewidth,keepaspectratio=false]{figures/branching_images/MM-1389-e.jpeg} \\
{\scriptsize\textbf{Figure A}} & {\scriptsize\textbf{Figure B}} &
{\scriptsize\textbf{Figure C}} & {\scriptsize\textbf{Figure D}} &
{\scriptsize\textbf{Figure E}} \\
\end{tabular}
\end{minipage}}}

\vspace{6pt}

{\setlength{\fboxsep}{6pt}%
\fcolorbox{prefixrule}{prefixbg}{%
\begin{minipage}{0.96\linewidth}
\footnotesize
\textbf{Prefix} (perceptual distractor; describes FSGS-like features,
inconsistent with PSGN), prepended to the model's response so the
watermark first acts at the next decision token:\\[2pt]
\textit{``In one of the glomeruli a portion of the tuft appears
collapsed and adherent to Bowman's capsule, while adjacent segments
look relatively preserved.''}
\end{minipage}}}

\vspace{6pt}

{\footnotesize
\textit{Sampled model continuations from this prefix} (verbatim, abridged):

\smallskip
\setlist[itemize]{nosep,leftmargin=1.4em,topsep=2pt,parsep=3pt,itemsep=3pt}
\color{black!55}
\begin{itemize}
\item \textit{Unwatermarked}: \emph{``$\ldots$\,This is consistent
  with FSGS, but the clinical scenario (recent streptococcal infection,
  facial oedema, haematuria with RBC casts) is classic for
  post-streptococcal glomerulonephritis$\ldots$''}
  \textcolor{black}{\textbf{$\Rightarrow$ A \,\textcolor{green!55!black}{\cmark}}}

\item \textit{Unwatermarked}: \emph{``$\ldots$\,The glomerulus has
  a ``lumpy-bumpy'' appearance from subepithelial immune complex
  deposition, characteristic of post-infectious glomerulonephritis
  $\ldots$''}
  \textcolor{black}{\textbf{$\Rightarrow$ A \,\textcolor{green!55!black}{\cmark}}}

\item \textit{PPL $\varepsilon{=}0.5$}: \emph{``$\ldots$\,The
  glomerulus shows segmental sclerosis with adhesion to Bowman's
  capsule, characteristic for focal and segmental
  glomerulosclerosis$\ldots$''}
  \textcolor{black}{\textbf{$\Rightarrow$ E \,\textcolor{red!75!black}{\xmark}}}

\item \textit{PPL $\varepsilon{=}0.5$}: \emph{``$\ldots$\,Bowman's
  capsule is distorted and the glomerular tuft appears partially
  collapsed; this is typical of focal segmental injury$\ldots$''}
  \textcolor{black}{\textbf{$\Rightarrow$ B \,\textcolor{red!75!black}{\xmark}}}
\end{itemize}}

\caption{\textbf{Branching at the next decision token (Q2, PSGN).}
The prefix describes FSGS-like features that are inconsistent with the
correct diagnosis (PSGN). \emph{Unwatermarked}, the model recognises
the misdirection and overrides the visual cue using the clinical
context (recent streptococcal infection, oedema, haematuria),
returning the correct answer. Under \emph{PPL $\varepsilon{=}0.5$
watermarking}, the perturbed next-token distribution causes
the continuation to commit to the cue's wrong interpretation, and the
rest of the reasoning chain anchors on FSGS-like findings.}
\label{fig:branching_example}
\end{figure}

\paragraph{Watermarks amplify distractors and suppress rescues.}
We illustrate the effect on Q2 (PSGN, \cref{fig:branching_example}):
a distractor prefix describing focal glomerular pathology
(inconsistent with the correct diagnosis) drops unwatermarked
accuracy by only $-3.3$\,pp ($64.0\% \to 60.7\%$); the model
dismisses the misdirection.
Under PPL watermarking, the same prefix produces a $-17.2$\,pp drop
($60.7\% \to 43.5\%$, $p{<}0.001$) -- about five times the
unwatermarked effect.

\Cref{tab:branching} reports the full set of eight configurations --
the Q2 example above plus seven additional cells spanning both prefix
mechanisms, both major watermark schemes, and a KGW strength sweep on Q3.
The effect manifests in two directions:
\emph{distractor amplification} (Q2 and Q3, where watermarking turns a
benign or mildly harmful prefix into a large accuracy drop) and
\emph{rescue suppression} (Q1, where a clinical hint that lifts
unwatermarked accuracy by $+18$\,pp is mostly erased by the
watermark).
Critically, on Q1 the same rescue prefix loses nearly identical
accuracy under KGW $\delta{=}3.0$ ($-15.3$\,pp) and PPL
$\varepsilon{=}0.4$ ($-14.3$\,pp)---two structurally different
watermarking mechanisms (green-list soft bias vs.\ constrained argmax
sampling) disrupting the same reasoning trajectory by similar amounts,
ruling out a scheme-specific artefact.
On Q3, stronger watermarking hurts more: $\Delta$ goes from
$-16.2$\,pp at $\delta{=}1$ to $-24.6$\,pp at $\delta{=}3$,
consistent with the degradation patterns in
\cref{sec:reasoning_degradation}.

\begin{table}[!htbp]
\centering
\small
\setlength{\tabcolsep}{4pt}
\caption{%
  Hash-averaged branching-point accuracy ($N{=}200$ paired trials).
  $\Delta = (\text{prefix\,+\,WM}) - (\text{prefix})$ is the
  accuracy loss attributable to watermarking on the prefix-grounded
  condition; $p$ from a two-proportion $z$-test.
}
\label{tab:branching}
\begin{tabular}{@{}llrrrrr@{}}
\toprule
        &                                                  & \multicolumn{3}{c}{Accuracy (\%)} &           &         \\
        \cmidrule(lr){3-5}
Q       & Prefix, scheme                                       & no prefix & prefix    & +\,WM     & $\Delta$  & $p$     \\
\midrule
Q1      & rescue, KGW $\delta{=}3.0$                           & $35.3$    & $53.3$    & $38.0$    & $-15.3$   & $0.001$ \\
Q1      & rescue, PPL $\varepsilon{=}0.4$                      & $35.3$    & $53.3$    & $39.0$    & $-14.3$   & $0.002$ \\
\midrule
Q2      & perceptual distractor, PPL $\varepsilon{=}0.5$       & $64.0$    & $60.7$    & $43.5$    & $-17.2$   & ${<}.001$ \\
Q2      & inferential distractor, PPL $\varepsilon{=}0.5$      & $64.0$    & $64.0$    & $46.5$    & $-17.5$   & ${<}.001$ \\
\midrule
Q3      & perceptual distractor, PPL $\varepsilon{=}0.5$       & $30.2$    & $30.7$    & $16.5$    & $-14.2$   & ${<}.001$ \\
Q3      & perceptual distractor, KGW $\delta{=}1.0$            & $30.2$    & $30.7$    & $14.5$    & $-16.2$   & ${<}.001$ \\
Q3      & perceptual distractor, KGW $\delta{=}2.0$            & $30.2$    & $30.7$    & $10.2$    & $-20.5$   & ${<}.001$ \\
Q3      & perceptual distractor, KGW $\delta{=}3.0$            & $30.2$    & $30.7$    & $6.1$     & $-24.6$   & ${<}.001$ \\
\bottomrule
\end{tabular}
\end{table}

\paragraph{Takeaway.}
This is a mechanistic probe on three questions, not a population-level
study, so it does not show that watermarks always degrade clinical
reasoning.
What it does show is that watermarks can cause harm at a specific,
predictable locus: the token where the model chooses between
competing clinical interpretations.
At these branching points, watermarking both amplifies misleading cues
the model would normally dismiss and suppresses helpful cues the model
would normally exploit---and two structurally different watermark
schemes (KGW and PPL) produce strikingly similar effects.
This offers a concrete explanation for the diverse reasoning failures
reported in \cref{sec:reasoning_degradation}, and it highlights a
blind spot of aggregate-accuracy benchmarks: because such benchmarks
sample questions uniformly, they dilute the signal from the subset of
questions where watermarking happens to perturb a decision-critical
token.

\clearpage
\section{Prompts}
\subsection{Generation Prompts}
\label{app:prompts}

The following prompts are used to elicit chain-of-thought reasoning from all evaluated models.

\begin{promptbox2}[MedQA]
You are a medical expert. Answer the following clinical multiple-choice question.\\[4pt]
Explain your reasoning, including key clinical clues, then state your final answer in the following format:\\[4pt]
FINAL ANSWER: \textless A, B, C, or D\textgreater
\end{promptbox2}

\begin{promptbox2}[MedXpertQA-MM]
You are a medical expert. Answer the following clinical multiple-choice question.\\[4pt]
Use the provided image(s) and question text to reason through the answer.\\[4pt]
Explain your reasoning, including key clinical clues, then state your final answer in the following format:\\[4pt]
FINAL ANSWER: \textless A, B, C, D, or E\textgreater
\end{promptbox2}

\subsection{Judge Prompts}
\label{app:judge_prompts}

This section summarises the three LLM-as-a-judge audit protocols.
For each audit we provide (i)~the prompt structure and input variables, (ii)~the evaluation axes with decision rules, (iii)~the output schema, and (iv)~one calibration example.
Verbatim prompts with all calibration examples are released in the supplementary code repository.
For all audits, the judge is \textsc{Gemini-3-Flash-Preview} with temperature~0 and a 16{,}384-token output budget.
For each (model, watermark) configuration, audits are run at the lowest watermark strength achieving $\geq 99\%$ TPR at $1\%$~FPR.

\subsubsection{Single-Response Audit (Audits on MedQA and MedXpertQA-MM)}
\label{app:audit1_prompt}

The judge evaluates one model completion at a time, blind to both the watermark condition and the gold answer. The MedQA and MedXpertQA-MM variants share the same axis definitions; the multimodal variant adds an image-grounding axis.

\paragraph{Input variables.}

\begin{promptbox2}[Prompt input]
\textbf{Role:} Board-certified physician evaluating a single response.\\[2pt]
\textbf{Clinical Question:} \pvar{question\_stem}\\
\textbf{Answer Options:} A.\,\pvar{option\_a} \quad B.\,\pvar{option\_b} \quad C.\,\pvar{option\_c} \quad D.\,\pvar{option\_d} \quad \textit{[E.\,\pvar{option\_e} for MedXpertQA-MM]}\\
\textbf{Image:} \textit{[Provided via API; MedXpertQA-MM only]}\\
\textbf{Submitted Response:} \pvar{output\_text}\\[4pt]
The judge does \textbf{not} see the gold answer.
\end{promptbox2}

\paragraph{Evaluation axes.}
\cref{tab:audit1_axes} summarises the axes and key decision rules. The terminology axis uses a priority-ordered decision tree (Rule~0$\to$1$\to$2$\to$3$\to$4); the misapplied-real-term classification requires all four prongs of a strict test to pass (named entity, assertional use, categorical distinctness, reasoning commitment). Full definitions, co-firing rules, and prong-field constraints are in the verbatim prompts.

\begin{table}[h]
\centering
\caption{Single-response audit: evaluation axes and key decision rules.}
\label{tab:audit1_axes}
\small
\begin{tabular}{@{}p{3.1cm}p{8.5cm}@{}}
\toprule
\textbf{Axis} & \textbf{What it captures} \\
\midrule
Terminology corruption & Corrupted spelling, fabricated entities, misapplied real terms (four-prong test). Priority-ordered decision tree; image-grounding terms routed to image axis first (Rule~0, multimodal only). \\[3pt]
Image grounding \newline {\scriptsize (MedXpertQA-MM only)} & Each visual claim classified as \textsc{perceptual} or \textsc{inferential}; perceptual claims scored \textsc{supported}\,/\,\textsc{contradicted}\,/\,\textsc{unverifiable} against the image. \\[3pt]
Vignette fabrication & Clinical scenario incompatible with the question stem (different demographics, presenting features, lab values). \\[3pt]
Within-response contradiction & Two committed positive assertions about the same entity that cannot both be true. \\[3pt]
Confident error & Fluent reasoning built on a verifiably false premise (fabricated entity, hallucinated image finding, wrong mechanism) without hedging. \\[3pt]
Clue misread & Stem detail used with the wrong concrete value (number, unit, laterality, timing). \\[3pt]
Linguistic corruption & Non-Latin script, mid-sentence language switches, Unicode artifacts, casing errors within terms. \\[3pt]
Format / termination failure & Truncated output, non-standard answer format, or post-answer reasoning loop. \\
\bottomrule
\end{tabular}
\end{table}

\noindent\textit{Distinguishing corrupted spelling from linguistic corruption.} Both flags catch surface-form damage, but at different granularity. \texttt{corrupted\_spelling} is a \emph{per-term} terminology flag for recognisable typos or transpositions that remain in Latin-alphabet English (e.g.\ ``cholecitis'' $\to$ cholecystitis). \texttt{linguistic\_corruption} is a \emph{response-level} boolean that fires when the language stream itself breaks: non-Latin script mid-English (e.g.\ ``sclerotic\,\begin{CJK}{UTF8}{gbsn}变化\end{CJK}''), mid-sentence language switches, invisible Unicode garbage, or casing/line-break artifacts inside terms.

\paragraph{Output schema (abbreviated).}

\begin{schemabox}[Output JSON - single-response audit]
\{\\
\quad "abstention": \{ "is\_none": bool, "cause": str|null \},\\
\quad "terminology\_flags": [ \{ "span", "classification", "rationale", ... \} ],\\
\quad "image\_grounding": \{ "image\_features\_claimed": [ \{ "claim", \\
\quad\quad "claim\_type": "PERCEPTUAL"|"INFERENTIAL",\\
\quad\quad "classification": "SUPPORTED"|"CONTRADICTED"|"UNVERIFIABLE"|null \} ] \},\\
\quad "format\_termination\_failure": \{ "evidence", "subtype", "present": 0|1 \},\\
\quad "vignette\_fabrication": \{ "evidence", "present": 0|1 \},\\
\quad "within\_response\_contradiction": \{ ..., "present": 0|1 \},\\
\quad "confident\_error": \{ "false\_premise", "downstream\_commitment", "present": 0|1 \},\\
\quad "clue\_misread": \{ "evidence", "present": 0|1 \},\\
\quad "linguistic\_corruption": \{ "evidence", "present": 0|1 \},\\
\quad "error\_summary": \{ counts and per-axis 0/1 flags \}\\
\}
\end{schemabox}

\paragraph{Calibration example (MedXpertQA-MM): image hallucination with confident error.}

\begin{calibrationbox}[Calibration - single-response audit]
\textbf{Vignette:} What is the most probable diagnosis in a 12-year-old boy who presents with recurrent infections?\\
\textbf{Options:} A.~Chronic granulomatous disease \quad B.~Brucellosis \quad C.~Burkitt's lymphoma \quad D.~Neurofibromatosis \quad E.~Tuberculous lymphadenitis\\[4pt]
\textbf{Image:}\\[2pt]
\begin{center}
\includegraphics[width=0.35\textwidth]{src/calibration_task_A.jpeg}
\end{center}
\vspace{-4pt}
\textbf{Submitted response (abridged):} ``The image shows a 12-year-old boy with a large, prominent mass on the right side of his neck [\ldots] with skin discoloration and visible superficial vascular patterning (telangiectasias). This clinical presentation -- particularly the large, soft-tissue mass with associated overlying skin vascular changes and caf\'e-au-lait spots -- is highly suggestive of neurofibromatosis type~I (NF1). [\ldots] FINAL ANSWER:~D''\\[6pt]
\textbf{Expected judge output (key fields):}
\begin{itemize}[nosep, leftmargin=1.2em]
\item \texttt{image\_grounding}: ``caf\'e-au-lait spots'' $\to$ \textsc{contradicted} (no discrete flat brown macules visible); ``skin discoloration'' $\to$ \textsc{contradicted}; ``telangiectasias'' $\to$ \textsc{unverifiable} (borderline: distended veins vs.\ telangiectasias).
\item \texttt{confident\_error}: present\,=\,1. False premise: hallucinated caf\'e-au-lait spots and telangiectasias. Downstream: entire NF1 diagnosis built on fabricated visual findings.
\item \texttt{terminology\_flags}: empty (all named terms are real and used correctly; the error is perceptual, not terminological).
\end{itemize}
\end{calibrationbox}

\subsubsection{Pairwise Divergence Audit (Audits on MedQA and MedXpertQA-MM)}
\label{app:audit2_prompt}

The judge sees both the watermarked and unwatermarked completions for the same item, with response order randomised and condition labels stripped. The gold answer is revealed for correctness scoring only.

\paragraph{Input variables.}

\begin{promptbox2}[Prompt input]
\textbf{Role:} Board-certified physician comparing two responses (order randomised, labels hidden).\\[2pt]
\textbf{Clinical Vignette:} \pvar{question\_stem}\\
\textbf{Answer Options:} A.\,\pvar{option\_a} \;\ldots\; D.\,\pvar{option\_d} \quad \textit{[E.\,\pvar{option\_e} for MedXpertQA-MM]}\\
\textbf{True answer:} \pvar{gold\_answer} \quad\textit{(used for correctness configuration only)}\\
\textbf{Image:} \textit{[Provided via API; MedXpertQA-MM only]}\\
\textbf{Response~1:} \pvar{response\_1\_text}\quad Selected: \pvar{response\_1\_answer}\\
\textbf{Response~2:} \pvar{response\_2\_text}\quad Selected: \pvar{response\_2\_answer}
\end{promptbox2}

\paragraph{Evaluation axes.}

The judge scores two dimensions:

\begin{enumerate}[nosep, leftmargin=1.4em]
\item \textbf{Correctness configuration:} \textsc{both\_correct}, \textsc{both\_wrong\_same\_answer}, \textsc{both\_wrong\_diff\_answer}, \textsc{r1\_only\_correct}, \textsc{r2\_only\_correct}, or \textsc{one\_or\_both\_none}.
\item \textbf{Diagnostic interpretation divergence} (\textsc{none}\,/\,\textsc{minor}\,/\,\textsc{major}): whether the two responses make mutually exclusive claims about the same clinical data (e.g., different mechanism attributions, image findings, or value interpretations).
\end{enumerate}

\paragraph{Output schema (abbreviated).}

\begin{schemabox}[Output JSON - pairwise divergence audit]
\{\\
\quad "configuration": \{ "response\_1\_answer", "response\_2\_answer",\\
\quad\quad "correctness\_configuration": str \},\\
\quad "diagnostic\_interpretation\_divergence": \{\\
\quad\quad "r1\_reading": str|null, "r2\_reading": str|null,\\
\quad\quad "description": str|null, "level": "NONE"|"MINOR"|"MAJOR" \}\\
\}
\end{schemabox}

\paragraph{Calibration example (MedQA): major divergence, same correct answer.}

\begin{calibrationbox}[Calibration - pairwise divergence audit]
\textbf{Vignette:} A 7-year-old boy is brought to the emergency department because of sudden-onset abdominal pain. Three days ago, he was diagnosed with a urinary tract infection and treated with nitrofurantoin. His parents emigrated from Kenya. Exam shows splenomegaly and scleral icterus. Labs: Hb~9.8, MCV~88, reticulocytes~3.1\%, total bilirubin~3.8 (direct~0.6), haptoglobin~16 (N\,=\,41--165), LDH~179. Which of the following is the most likely underlying cause?\\
\textbf{Options:} A.~Enzyme deficiency in RBCs \quad B.~Defective RBC membrane proteins \quad C.~Defect in orotic acid metabolism \quad D.~Absent hemoglobin beta chain\\
\textbf{True answer:} A\\[4pt]
\textbf{Response~1 (abridged):} ``[\ldots] G6PD deficiency renders red blood cells unable to regenerate NADPH [\ldots] Heinz body formation and bite cell morphology [\ldots] FINAL ANSWER:~A''\\
\textbf{Response~2 (abridged):} ``[\ldots] The most likely diagnosis is hereditary spherocytosis due to defective spectrin or ankyrin in the red cell membrane. However, since the question asks about enzyme deficiency [\ldots] the best answer is~A. FINAL ANSWER:~A''\\[6pt]
\textbf{Expected judge output:}
\begin{itemize}[nosep, leftmargin=1.2em]
\item \texttt{correctness\_configuration}: \textsc{both\_correct}.
\item \texttt{diagnostic\_interpretation\_divergence}: \textsc{major}. R1 diagnoses G6PD deficiency (enzyme defect, oxidative hemolysis); R2 diagnoses hereditary spherocytosis (membrane defect, extravascular hemolysis) and selects~A as a forced choice. Mutually exclusive pathophysiology despite the same final letter.
\end{itemize}
\end{calibrationbox}

\subsubsection{Reasoning-Trace Audit (MedQA, Reasoning Models)}
\label{app:audit3_prompt}

This audit targets reasoning models with extended chain-of-thought. The judge sees the full \texttt{<thinking>} trace and the post-\texttt{</think>} final answer for two responses (A and B), with position randomised and condition labels stripped.

\paragraph{Input variables.}

\begin{promptbox2}[Prompt input]
\textbf{Role:} Board-certified physician comparing two reasoning-model responses.\\[2pt]
\textbf{Question:} \pvar{question\_stem}\\
\textbf{Options:} A.\,\pvar{option\_a} \;\ldots\; D.\,\pvar{option\_d}\\[2pt]
\textbf{Response~A:} Reasoning trace \pvar{thinking\_text\_a} + Final answer \pvar{answer\_text\_a} + Selected \pvar{answer\_a}\\
\textbf{Response~B:} Reasoning trace \pvar{thinking\_text\_b} + Final answer \pvar{answer\_text\_b} + Selected \pvar{answer\_b}\\[4pt]
\textbf{Gold answer} \pvar{gold\_answer} \textit{is revealed only for Block~2 (factual accuracy).}
\end{promptbox2}

\paragraph{Evaluation protocol.}
The audit uses a two-block protocol to insulate process-quality scoring from outcome knowledge:

\textit{Block~1 -- Process quality} (scored before the gold answer is revealed): reasoning efficiency, clinical reasoning coherence, answer--reasoning alignment, distractor engagement.

\textit{Block~2 -- Factual accuracy} (gold answer revealed): terminological precision, diagnostic accuracy.

Each dimension is scored on a symmetric $\{-2, -1, 0, +1, +2\}$ scale describing Response~B relative to Response~A. Full anchoring definitions are in the verbatim prompt.

The judge also assigns two absolute per-response \textbf{clinical-deployment ratings}, scored in a separate single-pass evaluation where each response is assessed independently against the standard of care:

\begin{itemize}[nosep, leftmargin=1.4em]
\item \textbf{Clinical safety.} \textsc{no\_harm\_risk}: no identifiable safety concerns in either reasoning trace or final answer. \textsc{low\_risk}: minor issues unlikely to affect clinical outcome (e.g.\ imprecise but not misleading terminology, reasoning detour that self-corrects). \textsc{harm\_risk\_present}: content that could cause patient harm or clinical misunderstanding if used without verification -- including wrong answer, dangerous omission, alignment failure between reasoning and answer, lexical corruption of drug or disease names, or correct final answer supported by faulty reasoning.
\item \textbf{Supervisory burden.} \textsc{none}: safe to use as-is. \textsc{minor\_review}: final answer is sound but the reasoning trace contains inaccuracies that require verification. \textsc{major\_revision}: final answer or core reasoning must be reworked before clinical use.
\end{itemize}

\noindent The judge additionally identifies a single primary degradation pattern (\textsc{circularity}, \textsc{lexical\_corruption}, \textsc{logical\_drift}, \textsc{alignment\_failure}, \textsc{verbosity}, or \textsc{none}).

\paragraph{Output schema (abbreviated).}

The six relative change dimensions and summary are produced by the pairwise comparison pass; the clinical-deployment ratings are produced by the separate per-response pass. Both are shown below in a single combined schema.

\begin{schemabox}[Output JSON - reasoning-trace audit (combined)]
\{\\
\quad "score\_justifications": \{\\
\quad\quad "reasoning\_efficiency": str, "clinical\_reasoning\_coherence": str,\\
\quad\quad "answer\_reasoning\_alignment": str, "distractor\_engagement": str,\\
\quad\quad "terminological\_precision": str, "diagnostic\_accuracy": str \},\\
\quad "scores": \{\\
\quad\quad "reasoning\_efficiency": $\langle$-2 to +2$\rangle$, "clinical\_reasoning\_coherence": $\langle$-2 to +2$\rangle$,\\
\quad\quad "answer\_reasoning\_alignment": $\langle$-2 to +2$\rangle$, "distractor\_engagement": $\langle$-2 to +2$\rangle$,\\
\quad\quad "terminological\_precision": $\langle$-2 to +2$\rangle$, "diagnostic\_accuracy": $\langle$-2 to +2$\rangle$ \},\\
\quad "summary": \{\\
\quad\quad "justification": str,\\
\quad\quad "overall\_quality\_change": "IMPROVED"|...|"SEVERELY\_DEGRADED",\\
\quad\quad "primary\_degradation\_pattern": "CIRCULARITY"|...|"NONE" \},\\
\quad "clinical\_deployment": \{ \hfill \textit{// separate per-response pass}\\
\quad\quad "clinical\_safety": \{\\
\quad\quad\quad "justification": str,\\
\quad\quad\quad "rating": "NO\_HARM\_RISK"|"LOW\_RISK"|"HARM\_RISK\_PRESENT" \},\\
\quad\quad "supervisory\_burden": "NONE"|"MINOR\_REVIEW"|"MAJOR\_REVISION" \}\\
\}
\end{schemabox}

\paragraph{Calibration example: severe degradation via lexical corruption.}

\begin{calibrationbox}[Calibration - reasoning-trace audit]
\textbf{Question:} 45-year-old man with progressive dysphagia to solids, 10\,kg weight loss, barium swallow shows irregular narrowing of the distal oesophagus. Most likely diagnosis?\\
\textbf{Options:} A.~Achalasia \quad B.~Oesophageal stricture \quad C.~Oesophageal adenocarcinoma \quad D.~Diffuse oesophageal spasm \quad \textbf{Gold:}~C\\[4pt]
\textbf{Response~A reasoning (abridged):} Progressive dysphagia with weight loss and irregular narrowing points to malignancy. Distal location favours adenocarcinoma. Excludes achalasia (bird-beak, smooth tapering), stricture (smooth concentric narrowing), spasm (corkscrew). Linear, single pass. Selects~C.\\[2pt]
\textbf{Response~B reasoning (abridged):} Uses ``adeno-carsonoma'' instead of adenocarcinoma. Says ``barretts metaplasma'' instead of Barrett's metaplasia. Writes ``dysphagia'' correctly then switches to ``dysphasia'' (a speech disorder). Initially selects~A (achalasia), re-reads barium findings, switches to~C. Multiple self-corrections. Selects~C.\\[6pt]
\textbf{Expected judge output (key fields):}
\begin{itemize}[nosep, leftmargin=1.2em]
\item \texttt{scores}: reasoning\_efficiency\,=\,$-2$, clinical\_reasoning\_coherence\,=\,$-1$, answer\_reasoning\_alignment\,=\,$-1$, distractor\_engagement\,=\,$0$, terminological\_precision\,=\,$-2$, diagnostic\_accuracy\,=\,$0$.
\item \texttt{overall\_quality\_change}: \textsc{severely\_degraded}.
\item \texttt{primary\_degradation\_pattern}: \textsc{lexical\_corruption}.
\item Three corrupted terms flagged: ``adeno-carsonoma'', ``metaplasma'', ``dysphasia''. Initial answer flip to~A before self-rescue.
\end{itemize}
\end{calibrationbox}

\fi


\end{document}